\documentclass[12pt]{article}
\usepackage{amsmath}
\usepackage{graphicx}
\usepackage{natbib}
\usepackage{svg}

\newcommand{\blind}{0}

\usepackage[T1]{fontenc}    
\usepackage{hyperref}       
\usepackage{url}            
\usepackage{booktabs}       
\usepackage{amsfonts}       
\usepackage{microtype}      
\usepackage{xcolor}         
\usepackage{changes}        
\usepackage{nicefrac}       
\usepackage{multirow}
\usepackage{array}
\usepackage{natbib}

\usepackage{amssymb}
\usepackage{amsthm}
\usepackage{mathtools}
\usepackage{bm}
\usepackage{graphicx}
\usepackage{tikz}
\usepackage[capitalize,noabbrev]{cleveref}
\usepackage[ruled]{algorithm}
\usepackage{algpseudocode}
\usepackage{wrapfig}
\usepackage[most]{tcolorbox}
\usepackage{subcaption}
\usepackage{adjustbox}

\newcommand{\Rb}{\mathbb{R}}

\newcommand{\argmin}{\mathop{\rm arg~min}\limits}
\newcolumntype{K}[1]{>{\centering\arraybackslash}p{#1}}

\newcommand\mU{\mathbf{U}}

\newcommand\mY{\mathbf{Y}}

\newcommand\mZ{\mathbf{Z}}

\newcommand\CC{\mathcal{C}}

\newcommand\bmTheta{\boldsymbol{\Theta}}

\algdef{SE}[SUBALG]{Indent}{EndIndent}{}{\algorithmicend\ }%
\algtext*{Indent}
\algtext*{EndIndent}

\begin{document}

\def\spacingset#1{\renewcommand{\baselinestretch}%
{#1}\small\normalsize} \spacingset{1}


\if0\blind
{
  \title{\bf Tree-Guided $L_1$-Convex Clustering}
  \author{Bingyuan Zhang \thanks{
    The author gratefully acknowledge
    \textit{the Japan Science and Technology Agency, Grant Number JPMJSP2138.}}\hspace{.2cm} \\
    Institute for Glycocore Research, Nagoya University\\
    and \\
    Yoshikazu Terada \thanks{
    The author gratefully acknowledge \textit{the Japan KAKENHI Grant (JP20K19756, JP20H00601, JP23K28045, and JP24K14855), and the MEXT Project for Seismology Toward Research Innovation with Data of Earthquakes (STAR-E, JPJ010217)}.}\hspace{.2cm} \\
    Graduate School of Engineering Science, Osaka University}
  \maketitle
} \fi

\if1\blind
{
  \bigskip
  \bigskip
  \bigskip
  \begin{center}
    {\LARGE\bf Tree-Guided $L_1$-Convex Clustering}
\end{center}
  \medskip
} \fi

\bigskip
\begin{abstract}
Convex clustering is a modern clustering framework that guarantees globally optimal solutions and performs comparably to other advanced clustering methods.
However, obtaining a complete dendrogram (clusterpath) for large-scale datasets remains computationally challenging due to the extensive costs associated with iterative optimization approaches. 
To address this limitation, 
we develop a novel convex clustering algorithm called Tree-Guided $L_1$-Convex Clustering (TGCC). 
We first focus on the fact that the loss function of $L_1$-convex clustering with tree-structured weights can be efficiently optimized using a dynamic programming approach.
We then develop an efficient cluster fusion algorithm that utilizes the tree structure of the weights 
to accelerate the optimization process and eliminate the issue of cluster splits commonly observed in convex clustering. 
By combining the dynamic programming approach with the cluster fusion algorithm, 
the TGCC algorithm achieves superior computational efficiency without sacrificing clustering performance.
Remarkably, our TGCC algorithm can construct a complete clusterpath for $10^6$ points in $\mathbb{R}^2$ within 15 seconds on a standard laptop without the need for parallel or distributed computing frameworks.
Moreover, we extend the TGCC algorithm to develop biclustering and sparse convex clustering algorithms. 
\end{abstract}

\noindent%
{\it Keywords:}  Dynamic Programming; Minimum Spanning Tree; Hierarchical clustering.
\vfill

\newpage
\spacingset{1.5} 

\section{Introduction}
\label{sec1}

Clustering is an unsupervised learning to discover a hidden group (cluster) structure behind data. 
As a modern clustering framework, 
convex clustering has attracted significant attention \citep{pelckmans2005convex,lindsten2011clustering, hocking2011clusterpath, scalable2021, sun2021convex, Daniel:CCMM, chakraborty2023biconvex, Non-convexExtension2024}.
Convex clustering can be viewed as a regularized $k$-means clustering, and interestingly, 
it has both features of agglomerative hierarchical clustering and $k$-means clustering.
Like classical hierarchical clustering methods,
convex clustering provides a hierarchical clustering structure (i.e., a dendrogram) as a clustering result.
As its name indicates, 
the loss function of convex clustering is convex, 
ensuring that a globally optimal solution can be obtained, 
thus avoiding the problem of local minima encountered in the $k$-means clustering.

Now, we briefly introduce the convex clustering framework. 
Let $\boldsymbol{y}_1, \dots, \boldsymbol{y}_n \in \mathbb{R}^p$ be $p$-dimensional data points. 
For $q\ge 1$, $L_q$-convex clustering solves the following problem:
\begin{equation}
\label{main}
    \min_{\boldsymbol{\theta}}\frac{1}{2}\sum_{i = 1}^n\|\boldsymbol{y}_i - \boldsymbol{\theta}_i\|_2^2 + \lambda \sum_{i\neq j}w_{ij} \|\boldsymbol{\theta}_i - \boldsymbol{\theta}_j\|_q,
\end{equation}
where $\lambda >0$ is a tuning parameter, $w_{ij}$ is the weight between the $i$th and $j$th objects, and
$\|\bm{y}\|_q = \big(\sum_{i=1}^p|y_i|^q \big)^{1/q}$ for $\bm{y}=(y_1,\dots,y_p)^T\in \Rb^p$.
Among the different variations,
$L_1$- and $L_2$-convex clusterings are the most common and have been extensively studied \citep{radchenko2017convex, zhang2021dynamic, scalable2021, sun2021convex}. 
The $L_1$-convex clustering is a special case of the general fused lasso problem \citep{tibshirani2011solution}, 
making it computationally simpler than $L_2$-convex clustering. 
However, the $L_2$-convex clustering is rotationally invariant, 
meaning that rotating the data matrix does not affect the clustering result.
Except for sparse clusterings (e.g., \cite{wang2018sparse}), 
most clustering methods using the Euclidean distance have rotational invariance; the $L_2$-convex clustering is more popular than the $L_1$-convex clustering.

\begin{figure}[t]
\centering
\includegraphics[width = \textwidth]{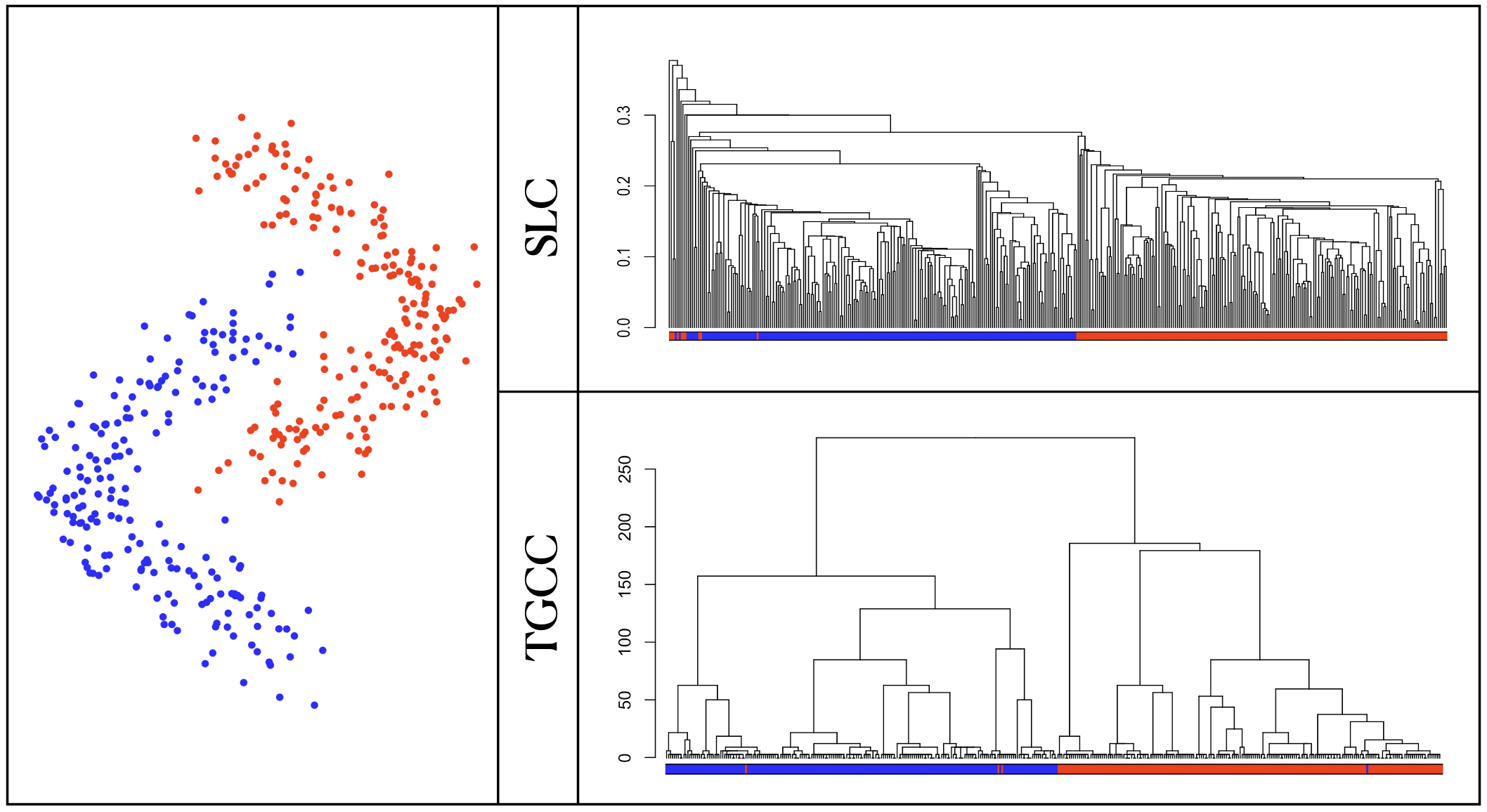}
\caption{A {\it Chaining} phenomenon of Single Linkage Clustering (SLC). 
Left: Two Moons data (TM).  
Right top: the result of SLC for the data in the left panel. 
Right bottom: the result of the proposed algorithm for the same data.}
\label{fig:intro}
\end{figure}

Once the optimal solution $\hat{\bmTheta}=(\hat{\boldsymbol{\theta}}_1,\dots,\hat{\boldsymbol{\theta}}_n)^T$ is obtained, 
the $i$th and $j$th objects are assigned to the same cluster if and only if $\hat{\boldsymbol{\theta}}_i$ = $\hat{\boldsymbol{\theta}}_j$.
With $\lambda = 0$, each object $\hat{\boldsymbol{\theta}}_i = \bm{y}_i$ becomes a distinct cluster.
As $\lambda$ increases, clusters gradually merge into larger clusters, and eventually, 
all the points merge into a single cluster for sufficiently large $\lambda$.
The continuous regularization path of solutions, called {\it clusterpath}, 
is obtained by solving the problem \eqref{main} for a sequence of increasing $\lambda$. 
The clusterpath can be visualized as a dendrogram (see the right bottom of Figure \ref{fig:intro}), 
allowing the number of clusters to be determined by cutting the dendrogram at an appropriate height.
%
%
However, there are several potential challenges in obtaining a convex clustering cluster path.
First, identifying clusters from the solution $\hat{\bmTheta}$ and 
solving the problem \eqref{main} for a sequence of $\lambda$s can be computationally prohibitive. 
Moreover, {\it cluster splits} may occur in the clusterpath. 
Cluster splits is a phenomenon in which objects fused at a lower $\lambda$ value split apart at a higher value of $\lambda$ \citep{hocking2011clusterpath}.
Since a larger value of $\lambda$ means stronger regularization enforcing fusion, 
cluster splits are an unnatural and unexpected phenomenon. 
Furthermore, cluster splits hinder the construction of a complete dendrogram and complicate the interpretation of clustering results \citep{chiquet2017fast}.


In practical applications of convex clustering, 
the choice of $w_{ij}$ is crucial for achieving good performance.
With proper weights, convex clustering methods yield superior performance \citep{sun2021convex}.
Whereas the identical weight case has been extensively studied \citep{tan2015statistical, radchenko2017convex, zhang2021dynamic},
convex clustering with identical weights often provides poor performance.
A common setting in practice is $w_{ij} = \exp(-\|\boldsymbol{y}_i-\boldsymbol{y}_j\|_2^2/\gamma)$ if $(i,j)\in E$ and zero otherwise,
where $E \subset \{1,\dots,n\}^2$ and $\gamma$ is a given constant.
The graph $E$ represents the edge structure of weights.
A $k$-nearest neighbor graph is commonly used as a weight structure $E$ for convex clustering. 
Weights with a sparse neighbor graph not only reduce the computational cost regarding the penalty terms but also guide the locally connected pairs to get fused, leading to better clustering performance \citep{hocking2011clusterpath, chi2015splitting, sun2021convex, Daniel:CCMM}.

On the other hand, geometric tree graphs play an important role in revealing the topological properties of data. 
For example, the Minimum Spanning Tree (MST) of a weighted complete graph is a tree that connects all the vertices while having the smallest total weights of the edges. 
It is known that the single linkage clustering (SCL) can be obtained from the MST \citep{gower1969minimum}. 
This equivalence enables the SLC to be solved in $\mathcal{O}(n^2)$-time and $\mathcal{O}(n)$-space (e.g., see \citet{murtagh1983survey}). 
However, the SLC has a well-known drawback called the problem of {\it chaining} (e.g., see Section 14.3.12 of \citet{ESL}).
In the SLC, we often see the {\it chaining} phenomenon in which data points are combined one by one into the previous (large) cluster in a chain-like fashion.
Then, the SLC tends to provide a very skewed hierarchy and straggly clusters that are hard to interpret.
In Figure~\ref{fig:intro}, the right top panel shows the dendrogram of the single linkage clustering for the Two Moons data shown in the left panel, and we see the chaining phenomenon of SLC occurs. 
Although the SLC gives poor clustering performance, $L_1$-convex clustering (the proposed method described later) with the corresponding MST provides a better clustering result (the right bottom of Figure \ref{fig:intro}).
The MST has been used in the clustering problem to improve clustering performance.  
\citet{yu2015hierarchical} proposed a new hierarchical clustering based on the geodesic distance of the MST. \citet{reddy2011mst} and \citet{yang2016minimum} use MST to improve the initialization for the $k$-means clustering. 

This paper develops a novel convex clustering algorithm called Tree-Guided $L_1$-Convex Clustering (TGCC), 
designed to construct a complete clusterpath and its corresponding dendrogram efficiently. 
We demonstrate that TGCC achieves a notable speed advantage over existing convex clustering algorithms without compromising performance through extensive experiments with both simulated and real data.
Remarkably, TGCC obtains a whole clusterpath in just about 60 seconds for $10^6$ points in $\mathbb{R}^2$, showing its practical applicability for large datasets. Moreover, since TGCC is the $L_1$-convex clustering, 
the optimization problem can be solved in parallel for each dimension\footnote{In all experiments of this paper, we do not use parallel computation for TGCC.}.
Thus, TGCC also has scalability for dimensions.

We summarize our contributions as follows:
\begin{itemize}
    \item We reveal that $L_1$-convex clustering with tree-structured weights can be solved efficiently by a dynamic programming algorithm for fused lasso.
    
    \item We propose a novel convex clustering algorithm called Tree-Guided $L_1$-Convex Clustering (TGCC) to construct a whole clusterpath with a hierarchical clustering result.
    In TGCC, the efficient cluster fusion step is essential for both the interpretability of the result and the acceleration of the algorithm.
    
    \item When we employ the MST corresponding to the single linkage clustering (SLC) as the structure of weights $w_{ij}$, 
    the TGCC can be viewed as an improvement of SLC. In fact, TGCC with the MST overcomes {\it the problem of chaining}.

    \item As important extensions of TGCC, 
    we also develop a biclustering algorithm and a sparse convex clustering algorithm based on TGCC. These algorithms demonstrate superior computational efficiency compared to existing methods.
\end{itemize}
The paper is organized as follows. Section \ref{sec2} introduces related works. Section \ref{sec3} presents the idea and the algorithms in detail. We analyze the time complexity and discuss the possibility of applying the TGCC in other settings. The experimental results are reported in Section \ref{sec4}. Finally, Section \ref{sec5} discusses the follow-up work. 

\section{Related work}
\label{sec2}

From an algorithmic perspective, developing efficient algorithms for convex clustering is an important research direction in the field.
\citet{lindsten2011clustering} proposed to use the CVX solver \citep{cvx} for solving the problem \eqref{main}. 
\citet{hocking2011clusterpath} proposed three different optimization methods for $\{1,2,\infty\}$-norms penalties. 
\citet{chi2015splitting} applied ADMM and AMA algorithms to solve convex clustering. 
Moreover, the second-order semi-smooth Newton method proposed by\citep{yuan2018efficient, sun2021convex} was reported to solve the problem \eqref{main} with a fixed value of $\lambda$, for $200,000$ data points in $\mathbb{R}^3$, within $6$ minutes.
Most recently, \citet{Daniel:CCMM} proposed an efficient majorization-minimization (MM) algorithm (CCMM) 
for $L_2$-convex clustering with general weights. 
The CCMM was reported to solve \eqref{main} with a fixed value of $\lambda$ with 50 seconds for $10^6$ data points in seven-dimensional space.
Most existing methods aim to develop an efficient algorithm for problem \eqref{main} with a single value of the tuning parameter $\lambda$. 
However, in order to obtain the whole clusterpath, we have to solve the problem for a sequence of $\lambda$, 
which is challenging and time-consuming. 
\citet{weylandt2020dynamic} proposed an approximation method called CARP to obtain the whole clusterpath adaptively based on ADMM. 
Although the optimization problem is generally difficult, efficient algorithms exist for special structures of weights. \citet{radchenko2017convex} and \citet{zhang2021dynamic} studied the particular case for $L_1$-convex clustering with identical weights and proposed scalable algorithms, but the setting is somewhat limited and less practical. 
\citet{chiquet2017fast} proposed another homotopy algorithm for the generalized fused-Lasso problem, but the structure of weights must be a complete graph, and the functional form for weights is restricted.

While the problem \eqref{main} is essentially difficult, efficient dynamic programming (DP) algorithms exist for the fused lasso problem \citep{tibshirani2005sparsity,tibshirani2011solution} when the structure of weights is a tree or a chain. 
\citet{johnson2013dynamic} first proposed a DP algorithm to solve the chain-structured fused lasso. When the input is one-dimensional, the time complexity is linear $\mathcal{O}(n)$ in sample size $n$. \citet{kolmogorov2016total} extended the idea to the tree-structured fused lasso and proposed an $\mathcal{O}(n\log n)$ algorithm using a complex data structure called Fibonacci Heap, which is hard to implement and is reported less efficient in practice \citep{padilla2017dfs}. \citet{kuthe2020engineering} suggested replacing the Fibonacci Heap with plain sorted queues, which results in a sub-optimal algorithm in time complexity: $\mathcal{O}(n^2\log n)$. However, the DP algorithm usually behaves $\mathcal{O}(n)$ in practice \citep{kuthe2020engineering}. 
As the tree-based DP method has been demonstrated to be highly efficient, 
it inspired us to apply the DP method to large-scale $L_1$-convex clustering. 

Theoretically, \citet{tan2015statistical} proved convex clustering with identical weights is closely related to SLC and $k$-means. \citet{pmlr-v70-panahi17a} showed that convex clustering with identical weights can recover clusters for cases such as Gaussian mixtures. \citet{sun2021convex} extended the result to the general weights setting, which is more practical. \citet{chiquet2017fast} studied the condition of weights that prevents the {\it cluster splits}. \citet{Non-convexExtension2024} provided sufficient conditions to recover the latent clusters for network lasso \citep{hallac2015network}, whose first term in loss \eqref{main} formulated in a more general form.

In addition, convex clustering can be extended into different problem settings.
\citet{chi2017convex} extended the convex clustering to the bi-clustering setting by introducing an additional penalty term in the loss function. 
They proposed a fast algorithm, called COBRA, to efficiently solve the optimization problem.
\citet{wang2018sparse} proposed the sparse convex clustering, which clusters observations 
and conducts feature selection simultaneously. 
\citet{supervised} proposed a new convex clustering algorithm using supervising auxiliary variables. 
\citet{chakraborty2023biconvex} proposed a method to optimize feature weights jointly and proposed a biconvex clustering method.
\citet{scalable2021} demonstrated a stochastic parallel algorithm for convex clustering.

\section{Method}
\label{sec3}

We propose to use tree-structured weights instead of $k$-NN structured weights in $L_1$-convex clustering. 
In our approach, we opt for tree-structured weights over the $k$-NN structured weights in convex clustering. 
These tree-structured weights allow us to leverage the efficient dynamic programming (DP) algorithm designed for the $L_1$-fused lasso problem, which is significantly more efficient than the iterative methods. 
Besides, the DP algorithm has another advantage that the DP solution is exact and does not suffer from numerical errors. 
Thus, we can obtain the clustering result by checking the {\it exact} equivalent relationships $\boldsymbol{\theta}_i = \boldsymbol{\theta}_j$ for any pair $\forall (i,j)$. In contrast, other iterative methods usually need a small threshold to check convergence conditions.

As pointed out in \cite{zhang2021dynamic}, $L_1$-convex clustering can be viewed as a direct application of the $L_1$-fused lasso problem.
For $L_1$-convex clustering, the loss function is separable across the features:
\begin{align}
    L(\bmTheta_n)
    &=\frac{1}{2}\sum_{i = 1}^n\|\boldsymbol{y}_i - \boldsymbol{\theta}_i\|_2^2 + \lambda \sum_{i<j}w_{ij} \|\boldsymbol{\theta}_i - \boldsymbol{\theta}_j\|_1 \nonumber\\
    &= \sum_{s = 1}^p \left\{ \frac{1}{2}\sum_{i = 1}^n(y_{is} - \theta_{is})^2 + \lambda \sum_{i\neq j}w_{ij} |\theta_{is} - \theta_{js}| \right\},
    \quad \bmTheta_n = (\theta_{is})_{n\times p} \in \Rb^{n\times p}.
    \label{eq:l1-cc}
\end{align}
Therefore, $L_1$-convex clustering simplifies into feature-wise 1-dimensional fused lasso problems. If the weights in the fused lasso problem have a special structure, we can solve the problem efficiently.
\cite{zhang2021dynamic} showed that $L_1$-convex clustering with identical and complete graph-structured weights could be converted into a weighted chain graph-structured fused lasso problem without heavy computations. 
This transformation enables the application of the DP algorithm \citep{johnson2013dynamic} for the fused lasso problem with chain-structured weights. However, as shown in the later numerical experiments, 
$L_1$-convex clustering with identical weights often performs poorly, 
even if the data has a clear cluster structure.

In this paper, we advocate for the use of tree-structured weights in convex clusteirng.
The Euclidean Minimum Spanning Tree (MST) is a prime example of the tree structure, 
as it contains all necessary information for performing single linkage clustering  \citep{gower1969minimum}.
The MST is closely related to clustering methods, 
and several clustering methods are based on the MST \citep{XuEtAl02, GrygorashEtAl06, ZhongEtAl11, WuEtAl13, yu2015hierarchical}.
To the best of our knowledge, 
integrating tree-structured weights with $L_1$ convex clustering is a novel approach.
By employing a highly efficient DP algorithm for tree-structured fused lasso problems, 
we can obtain the exact solution for a fixed $\lambda>0$.
To construct a complete clusterpath, we successively apply this algorithm to solve the problem \eqref{eq:l1-cc} for sequential values of $\lambda$.
However, this naive approach presents the following three issues:
\begin{itemize}
\item The computational cost increases with the number of $\lambda$ values required to construct a clusterpath.
\item For each $\lambda$, identifying clusters requires $O(n^2)$ comparisons, making the process computationally demanding.
\item The occurrence of {\it cluster splits} complicates the interpretation of clustering results. 
\end{itemize}

\subsection{Tree-Guided Convex Clustering (TGCC)}
\label{sub3.3}
To address these issues, we propose the Tree-Guided $L_1$-Convex Clustering (TGCC), 
a computationally efficient algorithm that is guaranteed to construct a non-split clusterpath. 
Thus, the clusterpath obtained by TGCC can be visualized as an interpretable dendrogram. 
The key idea is a natural constraint when solving the convex clustering problem for each $\lambda$. 
We will show that the optimization problem \eqref{eq:l1-cc} with such a constraint can be rewritten as an unconstrained general fused lasso problem. 
The constraint not only eliminates cluster splits but also simplifies the optimization procedure for a sequence of $\lambda$.

Let $\lambda_1<\lambda_2<\dots<\lambda_T$ be sequential values of the regularization parameter $\lambda$.
Let $(\hat{\bm{\theta}}_1(\lambda_1),\dots,\hat{\bm{\theta}}_n(\lambda_1))$ be the solution of (\ref{eq:l1-cc}).
For $t = 1,2,\dots,T-1$, at the $(t+1)$-th step,
we successively solve the optimization problems of (\ref{eq:l1-cc}) with the following constraints:
\[
\text{($\star$) }\;\bm{\theta}_i = \bm{\theta}_j
\text{ for any pair } (i,j) \text{ such that }
\hat{\bm{\theta}}_i(\lambda_{t}) = \hat{\bm{\theta}}_j(\lambda_{t}),
\]
where $(\hat{\bm{\theta}}_1(\lambda_{t}),\dots,\hat{\bm{\theta}}_{n}(\lambda_{t}))$ are the solution of the previous (i.e., $t$-th) constrainted optimization problem. 
As noted previously, checking whether the exact equality relationship $\hat{\bm{\theta}}_i(\lambda_{t}) = \hat{\bm{\theta}}_j(\lambda_{t})$ holds or not, 
we can obtain the clustering result at the $t$-th solution.
However, the computational cost of the naive comparison among all pairs is $\mathcal{O}(n^2)$ and is a computational bottleneck for large-scale data. 
Thus, we replace the constraint ($\star$) with the following relaxed constraint:
\[
\text{($\star^\prime$) }\;\bm{\theta}_i = \bm{\theta}_j
\text{ for any pair $(i,j)$ fused at the previous step}.
\]
Here, the constraint ($\star^\prime$) means
that the pairs $(i,j)$ never split in a later step if the $i$-th and $j$-th objects are assigned to the same cluster 
in the previous cluster fusion step.
In the next subsection, 
we will propose a tree-based cluster fusion algorithm to find the fused pairs in $\mathcal{O}(n)$ without checking all pairs. 

Under the constraint ($\star^\prime$), we first show that
the constrained optimization problem (\ref{eq:l1-cc})
can be rewritten as a smaller-size unconstrained optimization problem.
After the cluster fusion step, 
we obtain the partition (i.e., clustering result) $\CC = \{C_{1},\dots,C_{m}\}$ of the set $\{1,\dots,n\}$, where $m$ is the number of obtained clusters.
Since $\bm{\theta}_i =\bm{\theta}_j$ for any $i,j\in C_{k}$ under the constraint ($\star$),
each cluster can be represented by a single parameter vector.
Let $\boldsymbol{\theta}_k^{(c)}$ be the parameter vector corresponding to the $k$-th cluster,  
and  write $\bmTheta_m^{(c)} = (\bm{\theta}_1^{(c)},\dots,\bm{\theta}_{m}^{(c)})^T$.
Here, for $t=1$, each sample point forms its own cluster, and thus $\bmTheta_m^{(c)} =\bmTheta_n$.
Using these notations,
the constrained loss (\ref{eq:l1-cc}) can be rewritten as
\begin{equation}
\begin{aligned}
L\big(\bmTheta_m^{(c)} \big)
&=
\frac{1}{2}\sum_{k = 1}^{m}\tilde{\mu}_k
\|\tilde{\boldsymbol{y}}_k - \boldsymbol{\theta}_k^{(c)}\|_2^2 
+ 
\lambda \sum_{k=1}^{m}\sum_{l\neq k}^{m}\tilde{w}_{kl}
\|\boldsymbol{\theta}_k^{(c)} - \boldsymbol{\theta}_l^{(c)}\|_1
+
\frac{1}{2}\sum_{k = 1}^{m}\sum_{i \in C_k}\|\boldsymbol{y}_i-\tilde{\boldsymbol{y}}_k\|_2^2,
\label{eq:l1_cc_rewrite}
\end{aligned}
\end{equation}
where $\tilde{\mu}_k$ is the number of elements in the cluster $C_k$,
\begin{equation}
\label{update}
\tilde{\boldsymbol{y}}_k:=\frac{1}{\tilde{\mu}_k}\sum_{i\in C_k}\boldsymbol{y}_i,
\text{ and }
\tilde{w}_{kl}:=\sum_{i\in C_k}\sum_{j\in C_l}w_{ij}.
\end{equation}
Here, we note that the third term in \eqref{eq:l1_cc_rewrite} is a constant unrelated to the optimization problem. 
Thus, the problem \eqref{eq:l1_cc_rewrite} is a weighted version of the problem \eqref{eq:l1-cc}, and
this weighted problem can also be solved using the same optimization method.

Once we solve the smaller problem (\ref{eq:l1_cc_rewrite}), 
we get the parameter vectors $\hat{\bm{\theta}}_1^{(c)},\dots,\hat{\bm{\theta}}_m^{(c)}$ which represent the clusters 
obtained by the previous cluster fusion step.
Applying the cluster fusion algorithm to $\hat{\bmTheta}_m^{(c)} = (\hat{\bm{\theta}}_1^{(c)},\dots,\hat{\bm{\theta}}_m^{(c)})^T$, 
some clusters may be fused, and we obtain new clusters $\CC' = \{C_{1}',\dots,C_{m'}'\}\;(m'<m)$.
For the updated clusters $\CC'$, the loss function is also updated as \eqref{eq:l1_cc_rewrite} in the same manner.

If the updated weights $\tilde{w}_{kl}^{(t)}$ have a tree structure,
we can solve the unconstrained problem \eqref{eq:l1_cc_rewrite} efficiently by applying the DP algorithm.
However, there is no guarantee that the updated weights $\tilde{w}_{kl}^{(t)}$ are tree-structured.
For computational efficiency and interpretability, 
the following two desirable properties are required for a cluster fusion step:
\begin{itemize}
    \item[(C1)] Once two objects are assigned to the same cluster, they never split at later steps.
    \item[(C2)] The updated weights always have a tree structure at each step.
\end{itemize}
Here, the property (C1) is automatically satisfied under the constraint ($\star^\prime$), and 
our tree-based fusion algorithm described later ensures the property (C2). 

In the TGCC algorithm, 
the cluster path is constructed by alternately performing the DP algorithm to solve the problem \eqref{eq:l1_cc_rewrite} and the cluster fusion algorithm, proceeding sequentially from small values of $\lambda$.
Overall, the TGCC algorithm can be summarized as follows.

\begin{algorithm}[th]
\caption{\texttt{TGCC}}
\label{A1}
\begin{algorithmic}
    \State {\bfseries Input:}
    \State \quad Data matrix: $\mY = (\boldsymbol{y}_1,\dots,\boldsymbol{y}_n)^T$.
    \State \quad Increasing sequence of regularization parameters: $\lambda_1<\lambda_2<\dots<\lambda_T$.
    \State \quad Tree-structured weights: $\boldsymbol{w}=(w_{ij})$.
    \For{$t=1, \dots, T$} 
    \State{\bfseries Optimization step}: Solve problem \eqref{eq:l1_cc_rewrite} with $\lambda_t$ using the DP algorithm and obtain the optimal solution $\hat{{\bmTheta}}_m^{(c)}$.
    \State {\bfseries Cluster fusion step}:
    Apply the tree-based cluster fusion algorithm to $\hat{{\bmTheta}}_m^{(c)}$ and
    update $\tilde{\boldsymbol{y}}, \tilde{\boldsymbol{w}}$, and $\tilde{\boldsymbol{\mu}}$ by \eqref{update}.
    \EndFor
\end{algorithmic}
\end{algorithm}

\subsubsection{Tree-based cluster fusion algorithm} 
We propose a tree-based cluster fusion algorithm that satisfies the property (C2).
Let $\tilde{w}_{kl}^{(t)}$ be the updated weight between the $k$th and the $l$th clusters at the $t$-th iteration, as described above.
Assume the weights $\tilde{w}_{kl}$ have a tree structure,
and let $\mathcal{T}=(V,E)$ be the unweighted, undirected tree graph corresponding to these weights,
where $V =\{1,\dots,m\}$ is the set of all nodes, $m$ is the number of clusters at the current iteration, 
and $E$ is the set of all edges between nodes.
As described above, each node represents a cluster and corresponds to the parameter vector $\bm{\theta}_k^{(c)}$.
Let an arbitrary node be designated as the root of the tree.
Note that the choice of root does not affect the outcome of the fusion algorithm.
To satisfy the property (C2), 
we compare only pairs of nodes connected by an edge in the tree $\mathcal{T}$,
disregarding all other pairs in the proposed fusion algorithm.
The key point is ignoring the equivalence of pairs not in $E$, which ensures the property (C2).
By checking the equivalence $\hat{\bm{\theta}}_k^{(c)} = \hat{\bm{\theta}}_l^{(c)}$ for all edges $(k,l) \in E$,
we can obtain the updated clusters $\CC' = \{C_{1}',\dots,C_{m'}'\}\;(m'<m)$.
More precisely, we perform a tree traversal in breadth-first order, starting from the root, to check all edges.
Thus, the operation in our fusion algorithm is reduced from $\mathcal{O}(n^2)$ to $\mathcal{O}(n)$, and
the property (C2) is automatically satisfied.
The number of nodes decreases as the regularization parameter $\lambda$ increases 
while the updated weights remain tree-structured.

\begin{figure}[t]
    \centering
    \includegraphics[width=\textwidth]{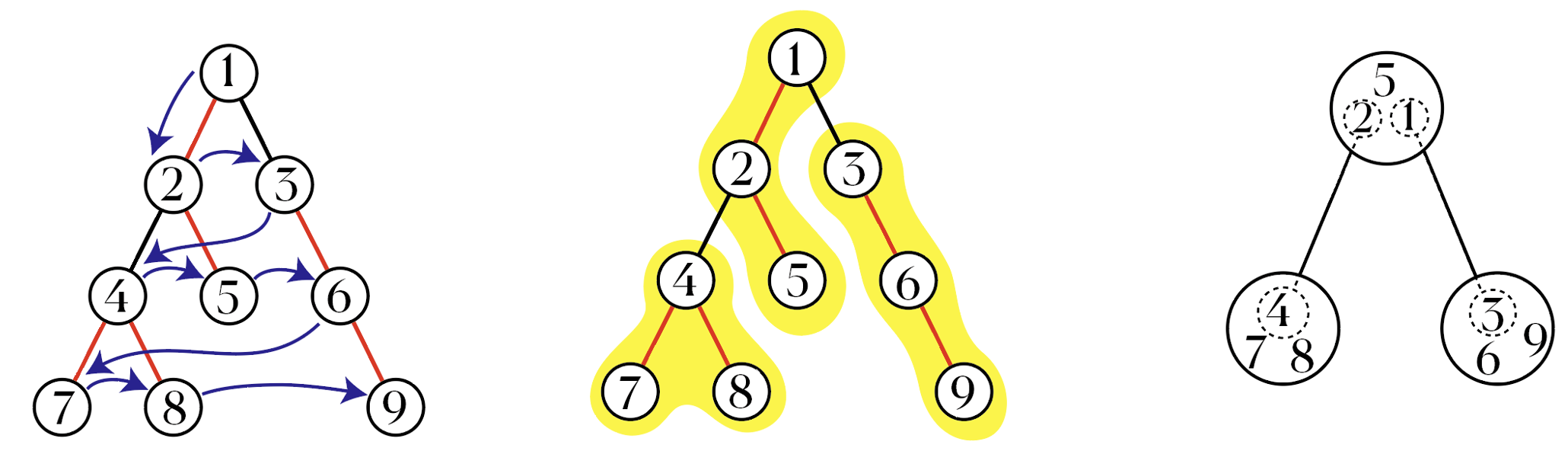}
    \caption{An illustration of the tree-based cluster fusion algorithm. Left: the original tree structure of nodes. Nodes sharing a red edge have the same $\hat{\boldsymbol{\theta}}$ values. Arrows: the tree traversal order via Breadth-First Search (BFS). Middle: the clusters found by the cluster fusion algorithm. Right: the new tree structure.}
    \label{fig:show_cluster_fusion_step}
\end{figure}

\subsection{Time complexity analysis}
\label{sub3.4}
To obtain a complete dendrogram, TGCC solves the problem \eqref{eq:l1_cc_rewrite} with a sequence of $(\lambda_1, \lambda_2, \dots, \lambda_T)$. 
The computational cost of the procedure includes two parts: the operations in the optimization step and the operations in the cluster fusion step. 
The convex clustering problem \eqref{eq:l1-cc} is solved by the dynamic programming method \citep{kuthe2020engineering}, 
whose time complexity is $\mathcal{O}(pn^2\log n)$. 
The cluster fusion algorithm searches through all edges, and a tree with $n$ nodes has $n-1$ edges. 
For each edge, 
the comparison between two parameter vectors $\boldsymbol{\theta}\in \mathbb{R}^p$ 
and the update of parameters in Algorithm~\ref{A1} takes $\mathcal{O}(p)$. 
Therefore, the time complexity of the cluster fusion algorithm is $\mathcal{O}(np)$. 
The total time complexity for TGCC is $\mathcal{O}(Tpn^2\log n)$. 
Though the time complexity is linear in the number of $\lambda$ and more than quadratic in the sample size $n$, 
we will numerically show later that the runtime is almost unaffected by the number of $\lambda$ 
and grows linearly as $n$ increases in practice.

\subsection{Biclustering and sparse convex clustering}

\label{sec3.3}
We extend the TGCC algorithm to develop biclustering and sparse clustering algorithms, called bi-TGCC and sp-TGCC, respectively.
We consider the following unified loss:
\begin{equation}
    L(\bmTheta) = \frac{1}{2}\|\mY - \bmTheta\|_F^2 + P_{1, \lambda}(\bmTheta) + P_{2, \gamma}(\bmTheta),
    \label{extgcc}
\end{equation}
where $\mY = (\boldsymbol{y}_1,\dots, \boldsymbol{y}_n)^T \in \Rb^{n\times p}$, 
and $\|\cdot\|_F$ denotes the Frobenius norm. 
Let $\bmTheta_{i\cdot}$ be the $i$-th row vector, and $\bmTheta_{\cdot i}$ be the $i$-th column vector of the matrix $\bmTheta$. 
The penalty terms $P_{1}$ and $P_{2}$ are defined as
\begin{equation*}
\begin{aligned}
    P_{1, \lambda}(\bmTheta) &= \lambda\sum_{i=1}^n\sum_{j\neq i}^n \alpha_{ij}\|\bmTheta_{i\cdot} - \bmTheta_{j\cdot}\|_1\quad\text{and}\\ 
    P_{2, \gamma}(\bmTheta) &= 
    \begin{cases}
        \gamma\sum_{i=1}^p\sum_{j\neq i}^p\beta_{ij}\|\bmTheta_{\cdot i} - \bmTheta_{\cdot j}\|_1, & \text{for biclustering},\\
        \gamma \sum_{i=1}^p \|\bmTheta_{.i}\|_2, &\text{for sparse clustering.}
    \end{cases}
\end{aligned}
\end{equation*}
The penalty $P_1$ is the penalty of $L_1$-convex clustering, and the penalty $P_2$ can be chosen accordingly for biclustering and sparse clustering. For bi-TGCC and sp-TGCC, we assume that the weights $\alpha_{ij}$ and $\beta_{ij}$ in the penalty terms are tree-structured. 
For bi-TGCC, $\alpha_{ij}, \beta_{ij}$ are computed using Gaussian kernels, 
with the tree structures of $\alpha_{ij}, \beta_{ij}$ being the minimum spanning trees for the rows and columns, respectively.

\begin{algorithm}[t]
\caption{\texttt{bi-TGCC/sp-TGCC}}
\begin{algorithmic}
\State {\bfseries Input:}{
Data matrix $\mY$, 
Weight $\boldsymbol{\alpha}$ in $P_1$, 
Weight $\boldsymbol{\beta}$ in $P_2$, 
Sequence $(\lambda_t, \gamma_t)_{t=1,\dots,T}$}\;
\State {\bfseries Output:}{ $(\widehat{\bmTheta}_1, \widehat{\bmTheta}_2, \dots, \widehat{\bmTheta}_T)$}\;
\State Set $\mU \gets \mY$.\;
\For{$t\gets 1,\dots,T$}
    \State $\mZ_1 \gets 0$, $\mZ_2 \gets 0$
    \Repeat
       \State $\mU' \gets \text{prox}_{P_{1,\lambda_t}}(\mU + \mZ_{1})$ 
       \State $\mZ_{1} \gets \mU + \mZ_{1} - \mU'$ 
       \State $\mU \gets \text{prox}_{P_{2,\gamma_t}}(\mU' + \mZ_{2})$ 
       \State $\mZ_{2} \gets \mU' + \mZ_{2} - \mU$ 
    \Until{converge.}
    \State $\hat{\bmTheta}_i \gets \mU$
    \State Perform the cluster fusion algorithm for the rows. 
    \State \bfseries{(For bi-TGCC)} Perform the cluster fusion algorithm for the columns. 
    \State \bfseries{(For sp-TGCC)} Perform the feature selection for the columns. 
    \EndFor
\end{algorithmic}
\label{alg:entend}
\end{algorithm}

Following \citep{chi2017convex}, 
a Dykstra-like proximal algorithm can be combined with the TGCC algorithm to solve the unified loss, 
and the obtained solution is guaranteed to converge to the unique global minimum of the objective \eqref{extgcc} \citep{chi2017convex, combettes2011proximal, tibshirani2017dykstra}. 
The unified algorithm for bi-TGCC and sp-TGCC is shown in Algorithm \eqref{alg:entend}. 
The proximal mapping of the function $P_{\lambda}$, $\text{prox}_{P_{\lambda}}(\mU)$ is defined as
\begin{equation*}
    \text{prox}_{P_\lambda}(\mU) = \argmin_{\bmTheta} \left[ \frac{1}{2}\| \mU - \bmTheta\|_F^2 + P_\lambda (\bmTheta) \right].
\end{equation*}

For bi-TGCC, solving $\text{prox}_{P_1, \lambda}(\cdot)$ and $\text{prox}_{P_2, \gamma}(\cdot)$ is equivalent to solving two independent convex clustering problems, 
which can be solved efficiently using the dynamic programming method \citet{kuthe2020engineering}. 
Moreover, we can apply the cluster fusion algorithm for both rows and columns. 
Consequently, the trees for rows and columns are reduced as tuning parameters $\lambda$ and $\gamma$ increase. 

For sp-TGCC, the $\text{prox}_{P_2, \gamma}(\cdot)$ becomes a group lasso problem \citet{yuan2006model},
which can be solved efficiently using standard optimization methods, such as block coordinate descent or proximal gradient method. 
This paper employs the Fast Iterative Shrinkage-Thresholding Algorithm (FISTA) \citet{beck2009fast} to solve the group lasso problem. 
To enhance the computational efficiency, we introduce a feature selection step to remove columns with zero values as $\gamma$ increases, which reduces the feature size. 

%

\section{Numerical Experiments}
\label{sec4}
In this section, we evaluate the computational efficiency and clustering performance of the proposed method 
relative to various existing clustering approaches. 
Numerical experiments were conducted with both synthetic and real datasets using an Apple M1 Pro chip.

We compare TGCC with other clustering methods, 
including existing convex clustering methods (CARP \citep{weylandt2020dynamic}, CPAINT \citep{zhang2021dynamic}, CCMM \citep{Daniel:CCMM}) and standard clustering methods (single linkage clustering (SLC), complete linkage clustering (CL), spectral clustering (SC; \citep{ng2001spectral}), DBSCAN (DBS; \citealp{ester1996density}), $k$-means (KM)). 
For the existing convex clustering methods, we used the source codes provided online by the authors.
Specifically, CARP, CCMM, and CPAINT are implemented in Rcpp. 
SCL and CL are available in the R package \texttt{stats}, 
and DBSCAN is implemented in the R package \texttt{dbscan}. We implemented the k-means++, spectral clustering, and the proposed method (TGCC, bi-TGCC, sp-TGCC) ourselves in Rcpp and R.
TGCC, bi-TGCC, and sp-TGCC are implemented in the R package \texttt{tgcc}, which is available at \url{https://github.com/bingyuan-zhang/tgcc}. All simulation codes used to generate the figures presented in the paper are available at \url{https://github.com/bingyuan-zhang/tgcc_simulation_code}. 

\subsection{Parameters setting}
For CARP, CCMM, and SSNAL, 
the weighted $k$-nearest neighbor ($k$-NN) graphs with the Gaussian kernel $w_{ij} = \exp(-\|\boldsymbol{y}_i-\boldsymbol{y}_j\|_2^2/\gamma)$ were used as the weights.
The number of neighbors $k$ was chosen to ensure connectivity among the samples in the datasets.
Note that CPAINT only supports uniform weights. 
For the Gaussian kernel, 
the parameter $\gamma$ was selected from predefined candidate sets.
For the $k$-NN structured weights, $\gamma$ was chosen from $\{0.5, 1, 2, 5, 10, 20, 50\}/\tau$. 
For the tree-structured weights, $\gamma$ was chosen from $\{1,2,5,10,20,50,100\}/\tau$, where 
\[
\tau := \frac{2}{n(n-1)}\sum_{i\neq j}\|\boldsymbol{y}_i - \boldsymbol{y}_j\|_2
\]
is a normalization factor. 
In the $k$-means clustering and the $k$-means step in the spectral clustering, 
we used the $k$-means++ \citep{kmeans++} to create 100 near optimal initial starts. 
DBSCAN has two hyperparameters: the minimum number of points (MinPts) was chosen from $\{3,5,10\}$, 
and the radius of neighbors (eps) was selected from $\{0.2,0.4,0.6,0.8,1,2,4\}/\tau$. 
For a fair comparison, 
we reported the accuracy using the best-performing hyperparameters for each method. 
The desired number of hard clustering methods was set equal to the number of clusters in true labels. 
The results of convex clustering methods across different $\gamma$ values are shown in the appendix. 

\subsection{Simulation studies for TGCC}
\subsubsection{Simulation settings}
We consider the following simulated datasets: 
Gaussian Mixture 1 (GM1), Gaussian Mixture 2 (GM2), and Two Circles (TC) as shown in \cref{fig:intro}; 
Two Moons (TM) as shown in \cref{fig:sim_data}. 
Details of the data generation procedures can be found in the appendix.

\begin{figure}[ht]
\centering
\includegraphics[width = 0.95\textwidth]{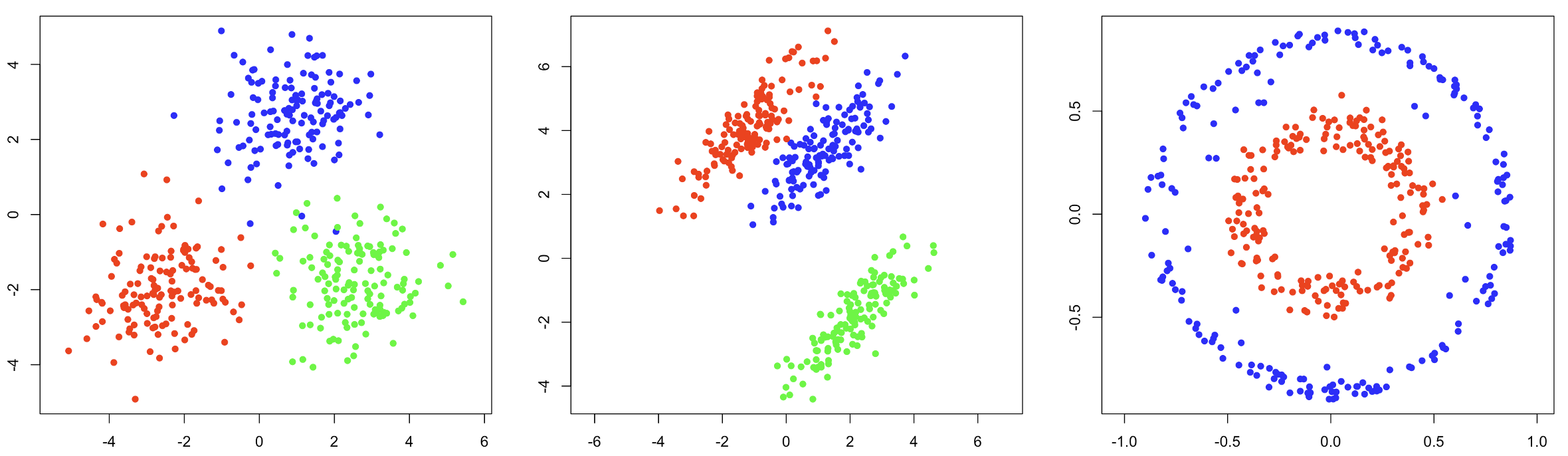}
\caption{Synthetic data (Left: Gaussian Mixture 1 (GM1), Middle: Gaussian Mixture 2 (GM2), Right: Two Circles (TC))}
\label{fig:sim_data}
\end{figure}

\subsubsection{Performance comparisons}

Table~\ref{table:sim} shows the clustering performance with a fixed sample size of $n=400$. 
Additional results for varying sample sizes are provided in the appendix. 
Clustering performance is evaluated based on classification accuracy, 
which is defined as the proportion of correctly predicted labels relative to the true labels. 
For convex clustering methods, 
the regularization parameter $\lambda$ was gradually increased until the desired number of clusters was obtained. 
The values reported in Table~\ref{table:sim} correspond to the median accuracy over 50 independent repetitions for each method.

\begin{table}[t]
\caption{Clustering performance (median accuracy) on simulated datasets over 50 repetitions.} 
\label{table:sim}
\centering
\begin{tabular}{c|ccccc|ccccc}
    \toprule
         Data & SLC & CL & SC & DBS & KM & CPAINT & CARP & CCMM & {\bf TGCC} \\
    \midrule
         GM1  & 0.340 & 0.966 & 0.988 & 0.828 & 0.988 & 0.728 & 0.985 & 0.988 & 0.984 \\
         GM2  & 0.668 & 0.769 & 0.988 & 0.891 & 0.888 & 0.690 & 0.988 & 0.995 & 0.993  \\
         TM   & 0.503 & 0.886 & 0.903 & 0.932 & 0.875 & 0.670 & 0.955 & 0.993 & 0.985\\
         TC   & 0.503 & 0.530 & 0.564 & 0.500 & 0.518 & 0.733 & 0.663 & 0.706 & 0.723 \\
    \bottomrule 
\end{tabular}
\end{table}

As shown in Table~\ref{table:sim}, 
both SLC and CPAINT failed to produce meaningful clusters, 
with their performance not exceeding that of random guesses.
The poor performance of SLC was attributed to the chaining problem.
The identical weights induce poor performance in CPAINT. 
Other convex clustering methods, including TGCC, showed comparable performance.
Although TGCC is restricted to using only tree-structured weights,
this limitation did not negatively affect its clustering performance. 
In fact, its results were comparable with those of other convex clustering methods utilizing $k$-NN structured weights.

\subsubsection{Runtime with increasing sample size}
Next, we evaluate the runtime of TGCC in comparison to other convex clustering methods. 
We generated the GM1 dataset with sample size $n$ ranging from $10^2$ to $10^6$ 
and reported the average runtime of five repetitions. 
\begin{figure}[ht]
    \centering
    \includegraphics[width=0.75\linewidth]{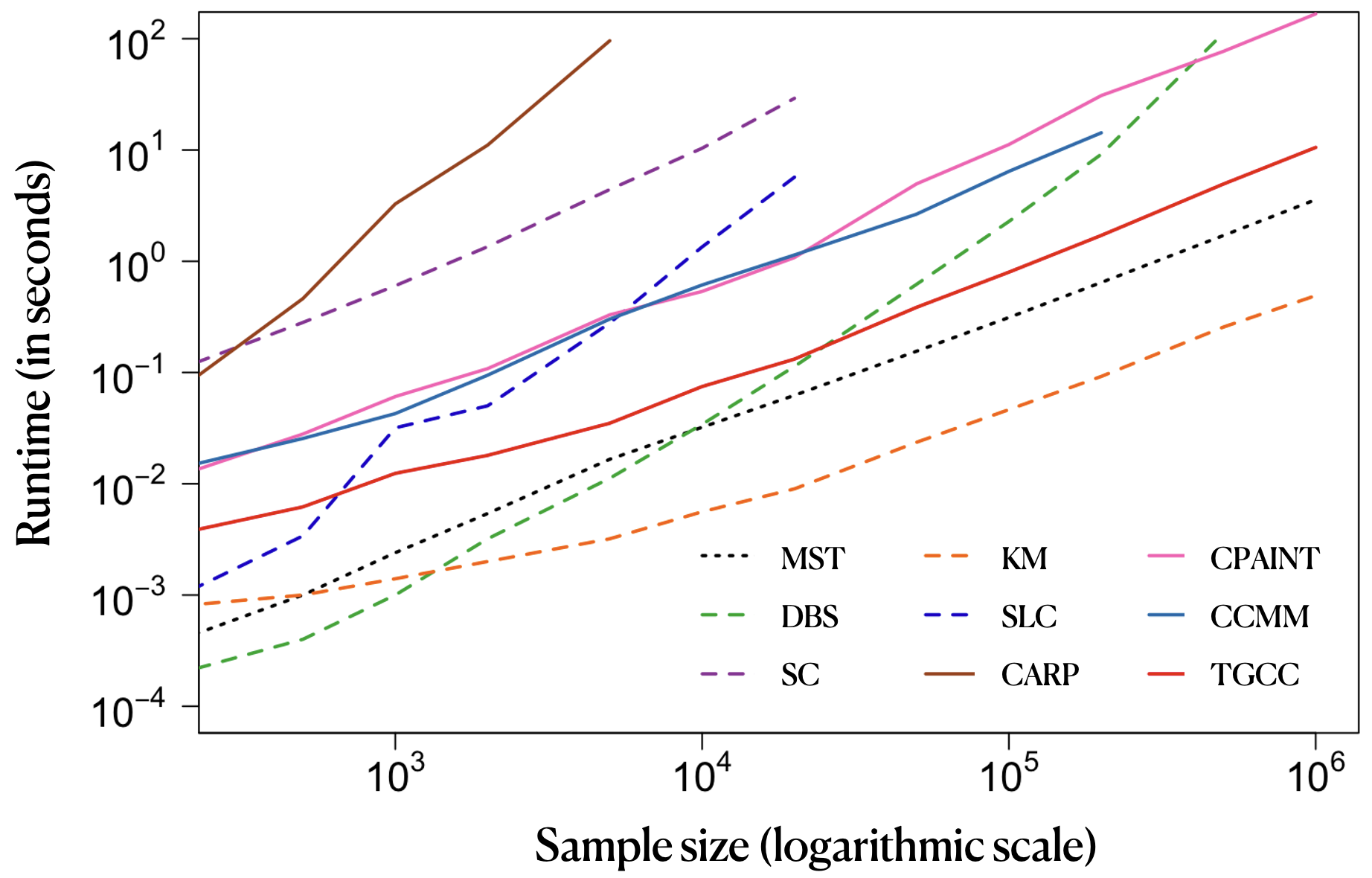}
    \caption{Runtime comparison of TGCC and other clustering algorithms. Average runtimes over five repetitions}
    \label{fig:runtime2methods}
\end{figure}

For the convex clustering methods, 
the reported runtimes correspond to the average real times spent in the optimization to construct the complete clusterpaths.
We utilized a sequence of 100 equally spaced values for the regularization parameter $\lambda$ for CPAINT, SNNAL, CCMM, and TGCC.
For CARP, which approximates the solution by gradually increasing $\lambda$ through multiplication by a small factor $t$, we set $t = 1.05$ as recommended in the original paper.
The runtimes for constructing the minimum spanning trees (MSTs) were recorded separately 
and have been included in the total runtimes of TGCC, as shown in Figure~\ref{fig:runtime2methods}. 
Additionally, the runtimes of standard clustering methods are reported for reference. 
For the $k$-means and spectral clustering,
we reported the runtime required to perform clustering with a single initial value.
Any method that exhausted the laptop’s memory (32 GB) or required more than 200 seconds was terminated early.

Figure~\ref{fig:runtime2methods} demonstrates that TGCC outperforms existing convex clustering algorithms in terms of runtime. The runtime of TGCC exhibits a linear growth pattern as the sample size $n$ increases, which can be attributed to the utilization of tree-structured weights. 
The cluster fusion algorithm is also a major contributing factor to this improvement. 

\subsubsection{Effect of the cluster fusion algorithm} 
\begin{figure}[ht]
    \centering
    \includegraphics[width=0.75\linewidth]{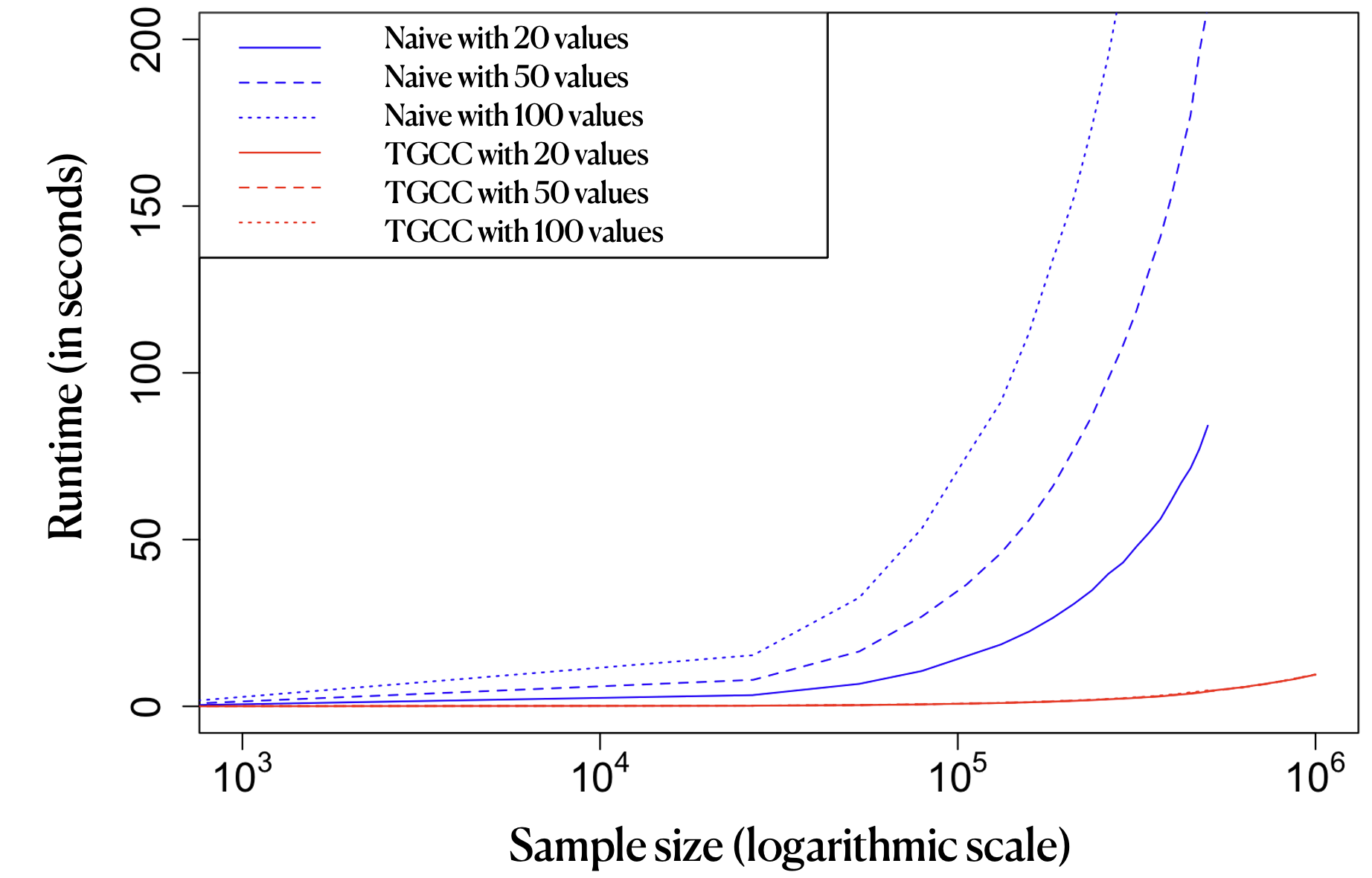}
    \caption{Runtime comparison of TGCC and the naive DP algorithm
            (Blue: the naive DP algorithm, Red: TGCC; 
            Solid: with 50 values of $\lambda$, Dashed: with 100 values of $\lambda$, Dotted: with 200 values of $\lambda$
            ).}
    \label{fig:runtime2dp}
\end{figure}

We evaluate the impact and effectiveness of the cluster fusion algorithm by comparing TGCC with a baseline method without the cluster fusion step.
The runtime of the baseline method increases substantially as the number of $\lambda$ values increases from $20, 50$ to $100$, while the runtime of TGCC remains unaffected (Figure~\ref{fig:runtime2dp}).
This property of TGCC provides a significant advantage in practical applications, 
as constructing a refined clusterpath (dendrogram) typically requires solving the convex clustering problem over a sequence of $\lambda$ values. 
Thus, the cluster fusion algorithm enables the TGCC to efficiently generate refined clusterpaths with low computational cost, even for large-scale datasets.

\begin{figure}
\centering
\includegraphics[width=0.75\linewidth]{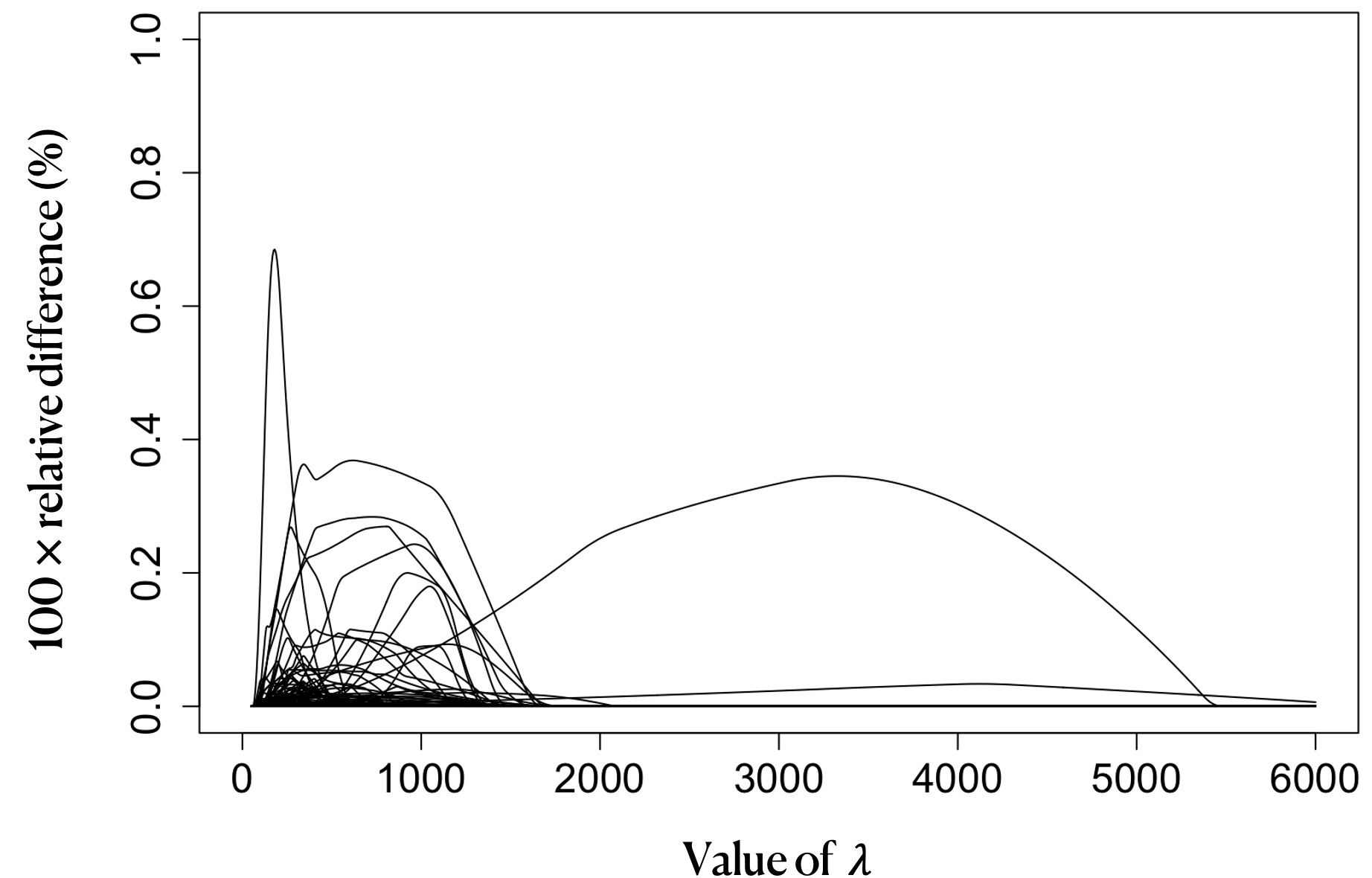} 
\caption{Effect of the cluster fusion algorithm in the optimality of the solution. Relative differences of 100 generated Gaussian Mixture 1 datasets ($n = 5000$).}
\label{fig:mergeeffect}
\end{figure}

More importantly, the cluster fusion algorithm has a negligible impact on the optimality of the solution to the underlying optimization problem.
To demonstrate this, we evaluate the relative difference between the loss at the TGCC solution and the loss at the true optimal solution for each value of $\lambda$. 
The relative difference (RD) is defined as 
\[
\text{RD}(\hat{\bmTheta}_{\text{tgcc}}(\lambda), \hat{\bmTheta}(\lambda)) 
:= \frac{L(\hat{\bmTheta}_{\text{tgcc}}(\lambda))  -  L(\hat{\bmTheta}(\lambda))}{L(\hat{\bmTheta}(\lambda))},
\]
where $\hat{\bm{\Theta}}_{\text{tgcc}}(\lambda)$ represents the solution obtained by the TGCC procedure at a given $\lambda$, and $\hat{\bm{\Theta}}(\lambda)$ denotes the optimal solution of the problem \eqref{eq:l1-cc} for the same $\lambda$.
We generated the Gaussian Mixture 1 dataset 100 times with a sample size of $n=5000$ and computed the RD-curve for each generated sample, as shown in Figure~\ref{fig:mergeeffect}.
The maximum relative difference observed across all 100 curves was less than $1\%$, demonstrating that the cluster fusion algorithm has minimal impact on the optimality of the solution.

\subsubsection{Effect of dimensions on runtime}
\begin{figure}[ht]
    \centering
    \includegraphics[width=0.45\linewidth]{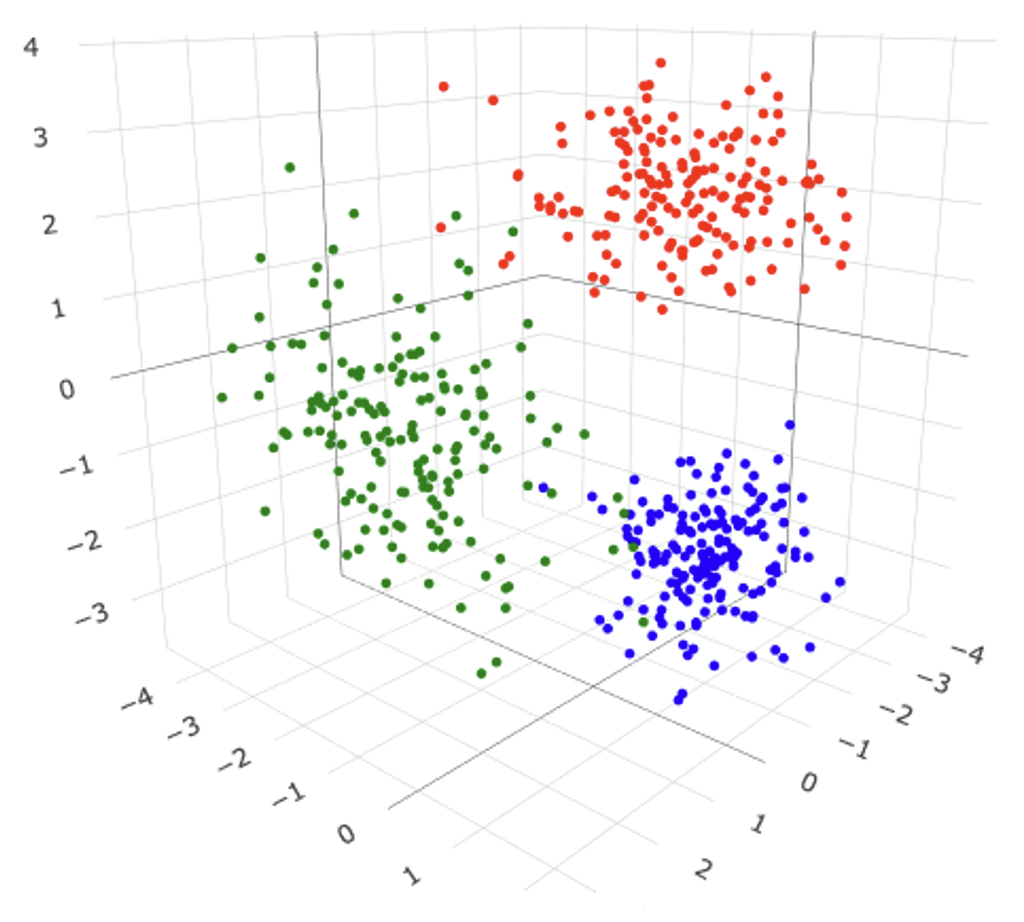}
    \caption{Synthetic data of Multidimensional Three Cluster (MTC).}
    \label{fig:MTC-model}
\end{figure}

Next, we compare the runtime of the TGCC fitting part and the calculation of MST with a fixed sample size and varied number of dimensions. In this simulation, 
we employ the simulation setup using the Multidimensional Three Cluster (MTC) model (Figure~\ref{fig:MTC-model}). MTC model is a generalized multidimensional version of the GM1. Specifically, we generate 2-dimensional mean vectors for the three clusters, 
$\mu_1 = (4/\sqrt{3}, 0), \, \mu_2 = (-2/\sqrt{3}, 2), \, \mu_3 = (-2/\sqrt{3}, -2)$
Then, these two-dimensional vectors are multiplied by a random orthogonal rotation matrix $A_i \in \mathbb{R}^{p \times 2},  i=1, 2, 3$ 
such that $A_iA_i^T = \boldsymbol{I}_p$. 
Then each element in $\mu_iA_i^T\in \Rb^p$ is added by a Gaussian noise from $\mathcal{N}(0, 0.25)$. 
The covariance matrices for the three clusters are generated similarly. 
First, a covariance matrix $\Sigma_B$ is calculated from a randomly generated matrix $B \in \mathbb{R}^{50 \times p}$, 
where $B_{ij} \sim \mathcal{N}(0, 1)$ for $i = 1, \dots, 50$ and $j = 1, \dots, p$. 
The final covariance matrices are set to be $\Sigma = \Sigma_B \otimes ll^T$, 
where $l \in \mathbb{R}^p$ is a $p$-dimensional vector with elements uniformly drawn from the interval $[0.5, 1.4]$. 

\begin{figure}[ht]
    \centering  \includegraphics[width=0.5\linewidth]{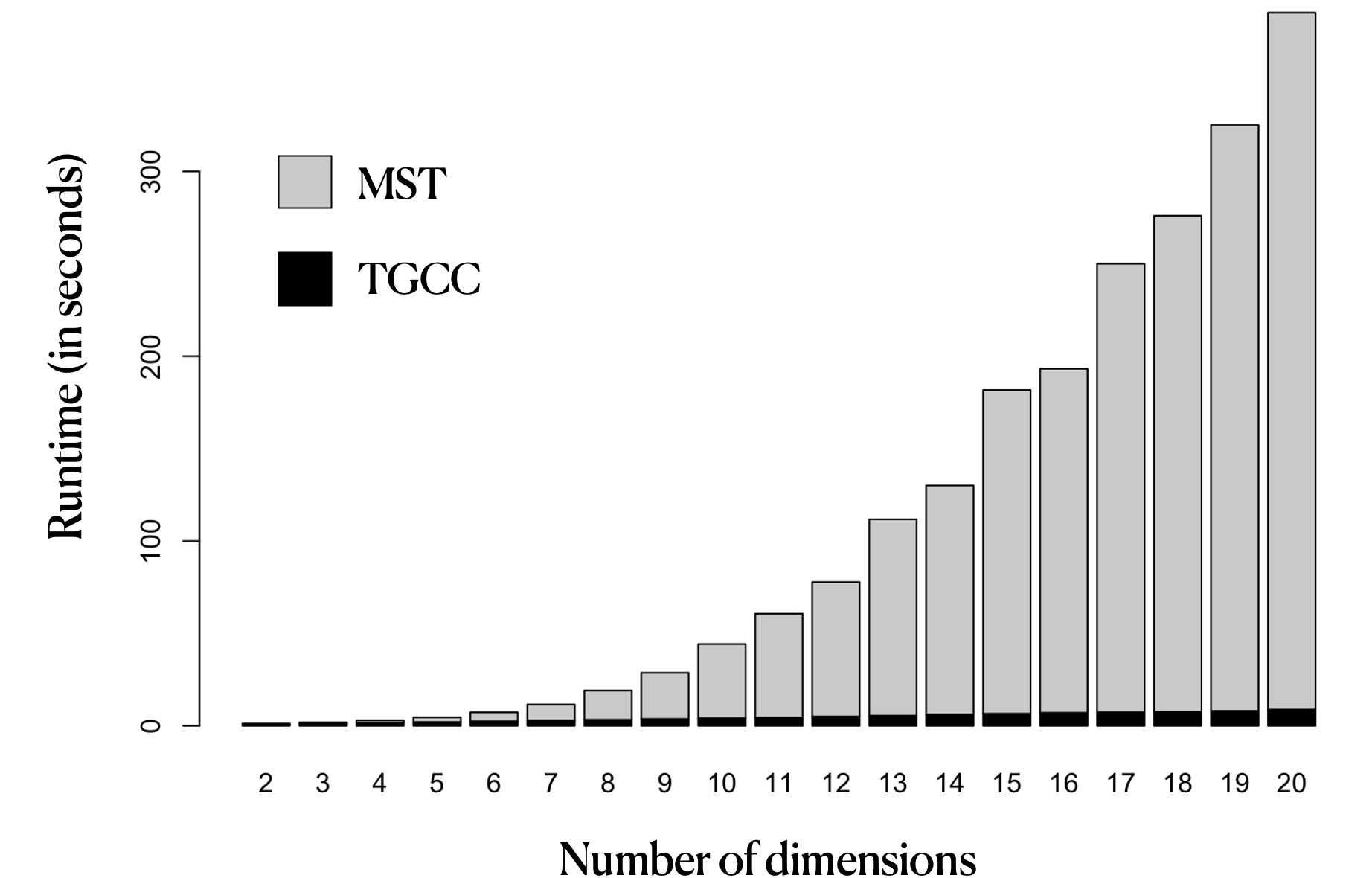}
    \caption{Runtime of TGCC (fitting) and MST with increasing number of dimensions and fixed sample size. Average runtimes over five repetitions.}
    \label{runtime2feature}
\end{figure}

The sample size is fixed to be $n=10^5$ (Figure~\ref{runtime2feature}), 
and the number of dimensions changes from 2 to 20. The mean runtimes over five independent repetitions are reported. The result shows the linearly increasing runtimes of TGCC fitting, which matches the previous analysis of the computational complexity. 
Importantly, it shows that the increase in dimensionality does not constitute a bottleneck for the optimization of TGCC. 
Compared to that, the computation cost of MST dominates the overall runtimes in TGCC as the number of dimensions increases. 
Although more efficient tree construction algorithms may be considered, it is beyond the scope of this paper. 

\subsection{Simulation studies for extensions of TGCC}
\subsubsection{Evaluation of bi-TGCC}

To demonstrate the performance of bi-TGCC, 
we compare it against the COBRA algorithm \citep{chi2017convex}, a state-of-the-art convex clustering method for bi-clustering. 
Following the simulation setting in \cite{chi2017convex}, 
we simulated the data matrices with a checkerboard bi-cluster structure. 
To generate a data matrix $\mY$, we first generate $y_{ij} \overset{i.i.d}{\sim} \, \mathcal{N}(\mu_{rc}, \sigma^2)$, 
and $\mu_{rc}$ takes one of the values equally spaced by $0.5$ between $-6$ and $6$. 
A low signal-to-noise ratio setting is considered, where the standard deviation is set to $\sigma=3$. 
The number of dimensions and sample size are set to be equal, with four row clusters and four column clusters.
Therefore, the resulting data matrix consists of $16$ bi-clusters. 
The data points are generated from one of the row groups randomly, 
and the probability of a row being assigned to the $i$-th group is inversely proportional to $i$. 
Similarly, the column groups are assigned in the same manner. 

\begin{figure}[ht]
    \centering
    \includegraphics[width=\linewidth]{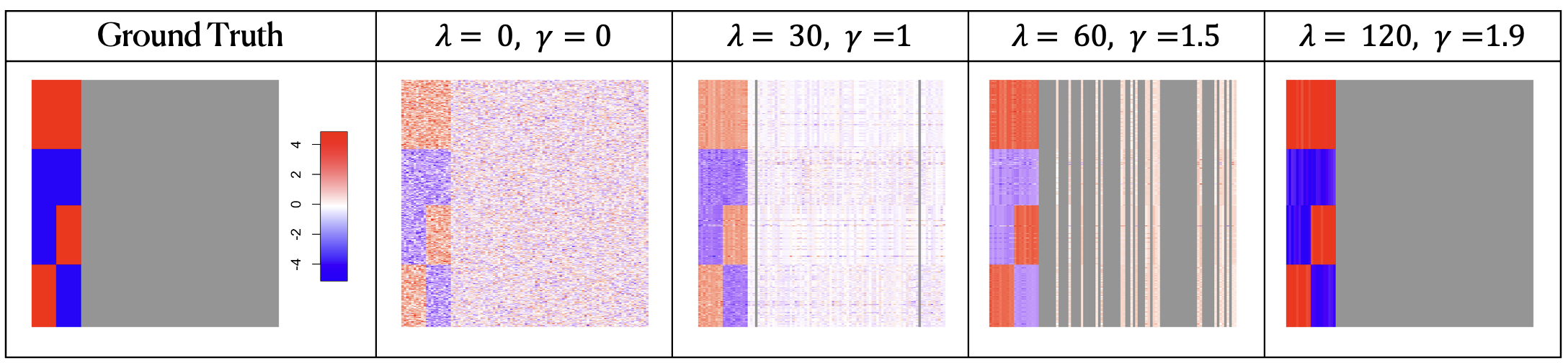}
    \caption{Results of the bi-TGCC algorithm.}
    \label{fig:bicc_ex}
\end{figure}

The left panel in Figure~\ref{fig:bicc_ex} shows the ground truth of the bi-clustering structures. 
The second panel shows an observed data matrix. 
As the tuning parameters increase, the bi-TGCC algorithm successfully recovers the underlying structures for both rows and columns. Table~\ref{table:extension1} shows the averages of the ARI, AC, and runtimes over ten repetitions. 
The same sequences of the tuning parameter and the same convergence criterion were used for bi-TGCC and COBRA to ensure a fair comparison. 
The sample sizes vary from 100 to 500. 
Compared to COBRA, the bi-TGCC algorithm is significantly faster while achieving comparable performance.

\begin{table}[ht]
\caption{Comparison of clustering accuracy and runtime between bi-TGCC and COBRA.}  
\label{table:extension1}
\centering
\begin{tabular}{lccccccc}
\toprule
     Setting & Methods & \multicolumn{2}{c}{AC} & \multicolumn{2}{c}{ARI} & \multicolumn{2}{c}{Runtime (s)} \\
\midrule 
\midrule 
     \multicolumn{2}{c}{} & Mean & SD & Mean & SD & Mean & SD \\
\midrule  
    $n=100$ & COBRA   & 0.840 & 0.092 & 0.680 & 0.155 & 6.232  & 5.066 \\ 
            & bi-TGCC & 0.993 & 0.016 & 0.981 & 0.043 & 0.308  & 0.198 \\
    $n=200$ & COBRA   & 0.914 & 0.064 & 0.809 & 0.130 & 8.923  & 8.755 \\
            & bi-TGCC & 0.934 & 0.080 & 0.897 & 0.111 & 0.857  & 0.201 \\
    $n=300$ & COBRA   & 0.902 & 0.084 & 0.804 & 0.138 & 39.855 & 49.914 \\
            & bi-TGCC & 0.982 & 0.056 & 0.968 & 0.102 & 1.872  & 0.033 \\
    $n=400$ & COBRA   & 0.924 & 0.029 & 0.836 & 0.069 & 55.285 & 15.476 \\
            & bi-TGCC & 0.993 & 0.022 & 0.988 & 0.038 & 3.592  & 0.087 \\
    $n=500$ & COBRA   & 0.947 & 0.023 & 0.873 & 0.058 & 131.363& 50.313 \\
            & bi-TGCC & 0.988 & 0.039 & 0.984 & 0.050 & 5.986  & 0.189 \\
\bottomrule
\end{tabular}
\end{table}

\subsubsection{Evaluation of sp-TGCC}
Next, we compare the sp-TGCC with the S-AMA optimization algorithm \citep{wang2018sparse} (SPCC) for spare convex clustering. 
The Four Spherical (FS) model \citep{wang2018sparse} is used to generate the data matrices. 
Data points are independently generated from a mixture of four multivariate normal (MVN) distributions. 
The first 20 informative features are generated from $\mathbf{MVN}(\boldsymbol{\mu}_k, \boldsymbol{I}_{20})$, $k \in \{1,2,3,4\}$. 
The mean vector $\bm{\mu}_k$ is defined as follows:
\begin{equation*}
    \begin{aligned}
        \bm{\mu}_k &= 
        (\mu\bm{1}_{10}^T, \mu\bm{1}_{10}^T)^T I(k=1) + 
        (\mu\bm{1}_{10}^T, -\mu\bm{1}_{10}^T)^T I(k=2)\\
        &\quad +
        (-\mu\bm{1}_{10}^T, \mu\bm{1}_{10}^T)^T I(k=3) + 
        (-\mu\bm{1}_{10}^T, -\mu\bm{1}_{10}^T)^T I(k=4).
    \end{aligned}
\end{equation*}
Finally, the remaining $p-20$ noisy features are generated independently from the standard normal distribution $\mathcal{N}(0, 1)$.
We set $\mu = 1.5$ and vary the sample size from 100 to 500 with the feature size $p=100$.

\begin{figure}[t]
    \centering
    \includegraphics[width=\linewidth]{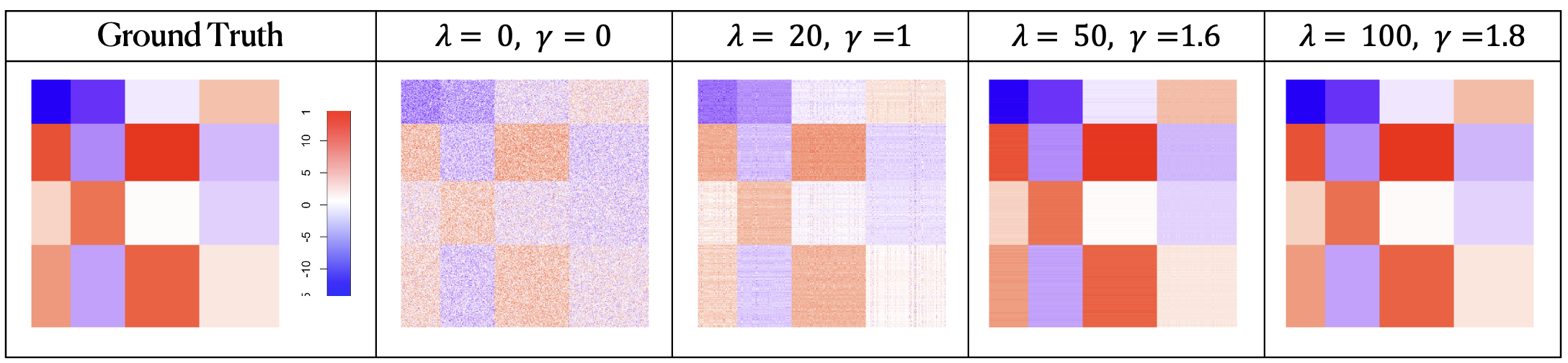}
    \caption{The results of teh sp-TGCC algorithm.}
    \label{fig:spcc_ex}
\end{figure}

The left panel in Figure~\ref{fig:spcc_ex} shows the ground truth (that is, the matrix of true mean vectors underlying the data matrix), 
with samples as rows and features as columns. 
The first 20 columns are the informative features while the remaining gray columns are the noisy features. 
The second panel shows the observed data matrix. 
The sp-TGCC algorithm recovers the cluster structures in samples and simultaneously removes the noisy features. 
Table \ref{table:extension2} shows the averages of the ARIs, ACs and runtimes over ten repetitions. 
The same tuning parameters sequences and the same convergence conditions were used for both SPCC and sp-TGCC. 
These results show that the sp-TGCC algorithm achieves a comparable performance with superior computational efficiency.

\begin{table}[ht]
\caption{Comparison of clustering accuracy and runtime between sp-TGCC and SPCC.}  
\label{table:extension2}
\centering
\begin{tabular}{lccccccc}
\toprule
     Setting & Methods & \multicolumn{2}{c}{AC} & \multicolumn{2}{c}{ARI} & \multicolumn{2}{c}{Runtime (s)} \\
\midrule 
\midrule 
     \multicolumn{2}{c}{} & Mean & SD & Mean & SD & Mean & SD \\
\midrule  
    $n=100$ & SPCC    & 0.962 & 0.058 & 0.965 & 0.050 & 29.692 & 6.402 \\
            & sp-TGCC & 0.958 & 0.091 & 0.941 & 0.124 & 2.052  & 0.147 \\
    $n=200$ & SPCC    & 0.985 & 0.041 & 0.986 & 0.034 & 81.873 & 16.579 \\
            & sp-TGCC & 0.997 & 0.010 & 0.991 & 0.022 & 2.136 & 0.121 \\
    $n=300$ & SPCC    & 0.983 & 0.031 & 0.983 & 0.025 & 154.763 & 24.370 \\
            & sp-TGCC & 0.977 & 0.071 & 0.967 & 0.091 & 2.365 & 0.146 \\
    $n=400$ & SPCC    & 0.943 & 0.083 & 0.943 & 0.084 & 261.816 & 54.447 \\
            & sp-TGCC & 1.000 & 0.001 & 0.999 & 0.002 & 2.239 & 0.113\\
    $n=500$ & SPCC    & 0.934 & 0.065 & 0.935 & 0.061 & 296.711 & 108.506\\
            & sp-TGCC & 1.000 & 0.000 & 1.000 & 0.000 & 2.279 & 0.156 \\
\bottomrule
\end{tabular}
\end{table}

\subsection{Application to real data}
\label{sec:4.2}

We analyze several real datasets for illustration.
Table~\ref{table:realdata} shows the clustering performance results for some representative real datasets, comparing SC, KM, CCMM, and TGCC. Here, the symbol ``--'' denotes unobserved values due to memory or time constraints.
If obtaining a result with a single hyperparameter took over one hour, the evaluation was terminated.
We observe that TGCC shows competitive performance to other advanced methods.
Detailed descriptions of the datasets can be found in the appendix, along with comprehensive results for other methods.
For datasets exceeding $10^5$ data points, spectral clustering (SC) was omitted due to the limited memory capacity. 
For the Covtype dataset, constructing a complete clusterpath with a fixed hyperparameter $\gamma$ took more than an hour for CCMM. 

\begin{table}[th]
\centering
\caption{Clustering performance (accuracy) on real datasets.}  
\label{table:realdata}
\begin{tabular}{crrrrrrr}
\toprule
Data                 & $n$     & $p$  & $K$  & SC     & KM     & CCMM   & \textbf{TGCC} \\
\midrule
Wine                &   178    & 13  & 3   & 0.955  & 0.966  & 0.972  & 0.910 \\
Breast Cancer       &   499    & 9   & 2   & 0.955  & 0.929  & 0.949  & 0.920 \\
Segmentation (small)&  2310    & 8   & 7   & 0.519  & 0.555  & 0.581  & 0.535 \\
MNIST (small)       & 10000    & 10  & 10  & 0.921  & 0.920  & 0.877  & 0.918 \\
Pendigits           & 10992    & 16  & 10  & 0.893  & 0.715  & 0.714  & 0.840 \\
DryBeans            & 13611    & 16  & 7   & 0.807  & 0.798  & 0.551  & 0.737 \\
HEPMASS (small)     & 100000   & 5   & 2   & --     & 0.880  & 0.533  & 0.892 \\
Covtype             & 581012   & 11  & 7   & --     & 0.259  & --     & 0.488 \\
\bottomrule
\end{tabular}
\end{table}

\begin{figure}[ht]
\centering
\includegraphics[width=0.7\textwidth]{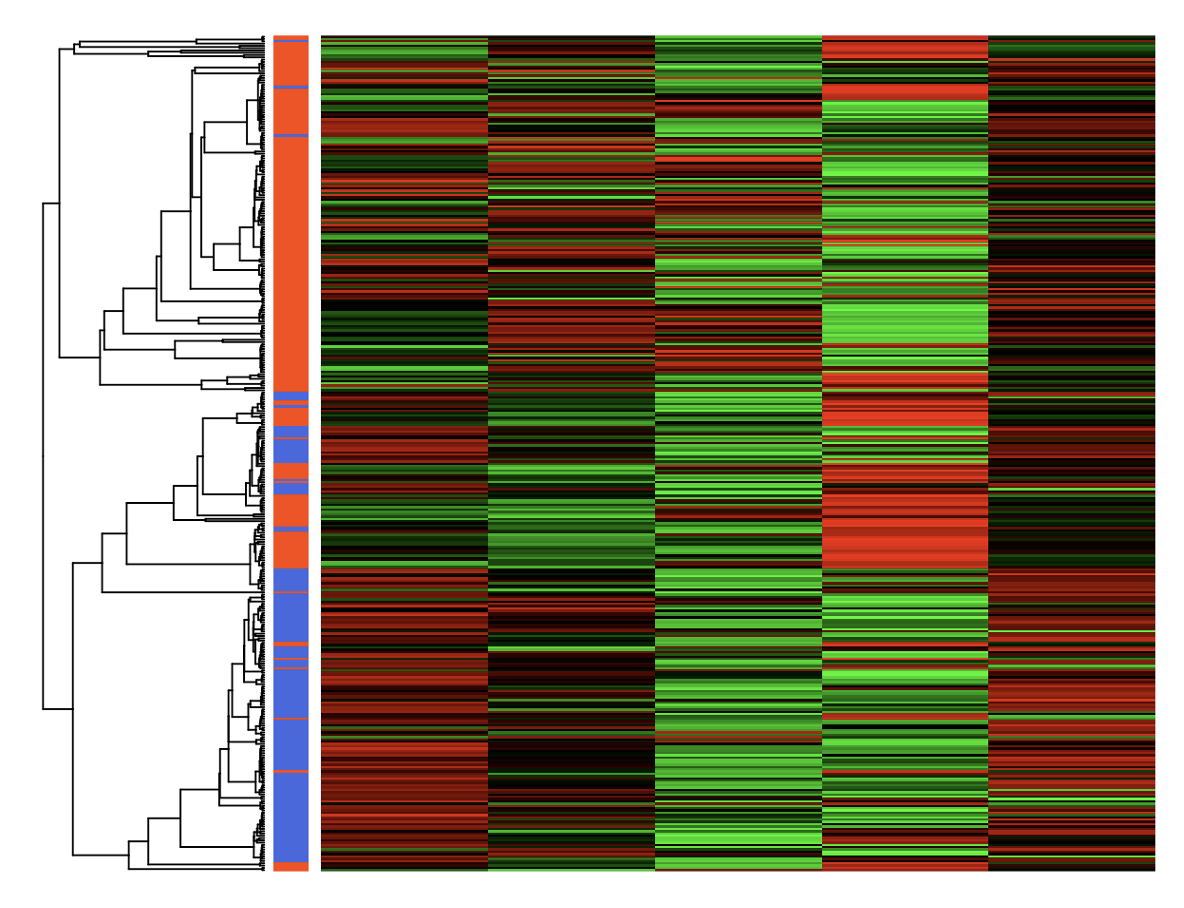}
\caption{The summarized heatmap of TGCC for HEPMASS data.}
\label{fig:heatmap}
\end{figure}

In addition, we apply the TGCC algorithm to a large-scale dataset.
We use a subset of the HEPMASS dataset, which contains data from millions of simulated particle collisions \citep{baldi2016parameterized}.
The dataset consists of two classes: particle signals and background noise.
From the test dataset in the HEPMASS data with mass = 1000,
we randomly extracted a sub-dataset with a sample size of $10^6$.
We then selected five relevant features out of the 27 normalized features to construct a 5-dimensional dataset with a sample size of $10^6$.
TGCC required approximately 80 seconds in total to construct the complete clusterpath for this dataset, of which 30 seconds were spent on the dynamic programming and cluster fusion algorithm.
The final clustering accuracy for the true labels is $88.93\%$.
Figure~\ref{fig:heatmap} displays the summarized heatmap with the estimated dendrogram. 
The rows correspond to the intermediate clusters produced by TGCC. 
The colors indicate the mean values of features in the intermediate clusters. 
The color bar on the left, positioned between the dendrogram and the heatmap, indicates the mode of labels in each intermediate cluster. 
Cutting the dendrogram into two clusters yields the clustering result of the TGCC algorithm. 
The dendrogram illustrates how similar signals are gradually merged into the final clusters. 
Consequently, the TGCC algorithm can produce meaningful hierarchical clustering results even for large-scale datasets within a reasonable computational time.

\section{Discussion}
\label{sec5}

In this paper, we proposed a novel convex clustering algorithm called Tree-Guided $L_1$-Convex Clustering (TGCC), 
which utilizes tree-structured weights and a dynamic programming approach to efficiently construct a complete clusterpath. 
The key component of the TGCC is the efficient cluster fusion algorithm, specifically designed to leverage the tree structure of the weights.
This cluster fusion algorithm not only addresses the computational challenges faced by existing convex clustering methods but also overcomes the issue of cluster splits, leading to a more interpretable hierarchical clustering structure.
Through extensive experiments on both synthetic and real-world datasets, 
we demonstrated that TGCC achieves superior computational efficiency without sacrificing clustering performance. 
Remarkably, TGCC can handle datasets with millions of data points in a few minutes on a standard laptop, 
making it applicable to large-scale clustering problems. 
Furthermore, we extended the TGCC framework to develop biclustering and sparse convex clustering algorithms. 

While we did not employ a parallel distributed algorithm (e.g., for MapReduce), 
the TGCC algorithm with a single core could be performed within approximately one minute for one million data points on a standard laptop.
Several large-scale clustering algorithms leverage parallel or distributed computing frameworks (e.g., \citealp{BateniEtAl17, MonathEtAl21, scalable2021}).
Developing a parallel or distributed version of the TGCC algorithm that can be efficiently executed on a large server is beyond the scope of this paper.
Based on the Massively Parallel MST algorithm, 
it may be possible to develop parallel distributed versions of the TGCC algorithm comparable to these large-scale clustering algorithms.

\noindent
{\bf Disclosure statement.} The authors report there are no competing interests to declare.

\bibliographystyle{plainnat}
\bibliography{Bibliography-MM-MC}

\begin{thebibliography}{}

\bibitem[\protect\citeauthoryear{Campbell and Austin}{Campbell and
  Austin}{2002}]{Campbell02}
Campbell, J.~I. and S.~Austin (2002).
\newblock Effects of response time deadlines on adults' strategy choices for
  simple addition.
\newblock {\em Memory \& Cognition\/}~{\em 30\/}(6), 988--994.

\bibitem[\protect\citeauthoryear{Chi, Feltovich, and Glaser}{Chi
  et~al.}{1981}]{Chi81}
Chi, M.~T., P.~J. Feltovich, and R.~Glaser (1981).
\newblock Categorization and representation of physics problems by experts and
  novices.
\newblock {\em Cognitive science\/}~{\em 5\/}(2), 121--152.

\bibitem[\protect\citeauthoryear{Schubert, Denmark, Crandall, Grome, and
  Pappas}{Schubert et~al.}{2013}]{Schubert13}
Schubert, C.~C., T.~K. Denmark, B.~Crandall, A.~Grome, and J.~Pappas (2013).
\newblock Characterizing novice-expert differences in macrocognition: an
  exploratory study of cognitive work in the emergency department.
\newblock {\em Annals of emergency medicine\/}~{\em 61\/}(1), 96--109.

\end{thebibliography}


\begin{thebibliography}{50}
\providecommand{\natexlab}[1]{#1}
\providecommand{\url}[1]{\texttt{#1}}
\expandafter\ifx\csname urlstyle\endcsname\relax
  \providecommand{\doi}[1]{doi: #1}\else
  \providecommand{\doi}{doi: \begingroup \urlstyle{rm}\Url}\fi

\bibitem[Arthur and Vassilvitskii(2007)]{kmeans++}
David Arthur and Sergei Vassilvitskii.
\newblock K-means++: The advantages of careful seeding.
\newblock In \emph{Proceedings of the Eighteenth Annual ACM-SIAM Symposium on Discrete Algorithms}, page 1027–1035, 2007.

\bibitem[Baldi et~al.(2016)Baldi, Cranmer, Faucett, Sadowski, and Whiteson]{baldi2016parameterized}
Pierre Baldi, Kyle Cranmer, Taylor Faucett, Peter Sadowski, and Daniel Whiteson.
\newblock Parameterized neural networks for high-energy physics.
\newblock \emph{The European Physical Journal C}, 76\penalty0 (5):\penalty0 1--7, 2016.

\bibitem[Bateni et~al.(2017)Bateni, Behnezhad, Derakhshan, Hajiaghayi, Kiveris, Lattanzi, and Mirrokni]{BateniEtAl17}
Mohammadhossein Bateni, Soheil Behnezhad, Mahsa Derakhshan, MohammadTaghi Hajiaghayi, Raimondas Kiveris, Silvio Lattanzi, and Vahab Mirrokni.
\newblock Affinity clustering: Hierarchical clustering at scale.
\newblock In \emph{Advances in Neural Information Processing Systems}, volume~30, 2017.

\bibitem[Beck and Teboulle(2009)]{beck2009fast}
Amir Beck and Marc Teboulle.
\newblock A fast iterative shrinkage-thresholding algorithm for linear inverse problems.
\newblock \emph{SIAM journal on imaging sciences}, 2\penalty0 (1):\penalty0 183--202, 2009.

\bibitem[Chakraborty and Xu(2023)]{chakraborty2023biconvex}
Saptarshi Chakraborty and Jason Xu.
\newblock Biconvex clustering.
\newblock \emph{Journal of Computational and Graphical Statistics}, pages 1--23, 2023.

\bibitem[Chi and Lange(2015)]{chi2015splitting}
Eric~C Chi and Kenneth Lange.
\newblock Splitting methods for convex clustering.
\newblock \emph{Journal of Computational and Graphical Statistics}, 24\penalty0 (4):\penalty0 994--1013, 2015.

\bibitem[Chi et~al.(2017)Chi, Allen, and Baraniuk]{chi2017convex}
Eric~C Chi, Genevera~I Allen, and Richard~G Baraniuk.
\newblock Convex biclustering.
\newblock \emph{Biometrics}, 73\penalty0 (1):\penalty0 10--19, 2017.

\bibitem[Chiquet et~al.(2017)Chiquet, Gutierrez, and Rigaill]{chiquet2017fast}
Julien Chiquet, Pierre Gutierrez, and Guillem Rigaill.
\newblock Fast tree inference with weighted fusion penalties.
\newblock \emph{Journal of Computational and Graphical Statistics}, 26\penalty0 (1):\penalty0 205--216, 2017.

\bibitem[Combettes and Pesquet(2011)]{combettes2011proximal}
Patrick~L Combettes and Jean-Christophe Pesquet.
\newblock Proximal splitting methods in signal processing.
\newblock \emph{Fixed-point algorithms for inverse problems in science and engineering}, pages 185--212, 2011.

\bibitem[Dua and Graff(2017)]{Dua:2019}
Dheeru Dua and Casey Graff.
\newblock {UCI} machine learning repository, 2017.
\newblock URL \url{http://archive.ics.uci.edu/ml}.

\bibitem[Ester et~al.(1996)Ester, Kriegel, Sander, and Xu]{ester1996density}
Martin Ester, Hans-Peter Kriegel, J\"{o}rg Sander, and Xiaowei Xu.
\newblock A density-based algorithm for discovering clusters in large spatial databases with noise.
\newblock In \emph{Proceedings of the Second International Conference on Knowledge Discovery and Data Mining}, pages 226--231, 1996.

\bibitem[Gagolewski et~al.(2016)Gagolewski, Bartoszuk, and Cena]{gagolewski2016genie}
Marek Gagolewski, Maciej Bartoszuk, and Anna Cena.
\newblock Genie: A new, fast, and outlier-resistant hierarchical clustering algorithm.
\newblock \emph{Information Sciences}, 363:\penalty0 8--23, 2016.

\bibitem[Gower and Ross(1969)]{gower1969minimum}
John~C Gower and Gavin~JS Ross.
\newblock Minimum spanning trees and single linkage cluster analysis.
\newblock \emph{Journal of the Royal Statistical Society: Series C (Applied Statistics)}, 18\penalty0 (1):\penalty0 54--64, 1969.

\bibitem[Grant and Boyd(2014)]{cvx}
Michael Grant and Stephen Boyd.
\newblock {CVX}: Matlab software for disciplined convex programming, version 2.1.
\newblock \url{http://cvxr.com/cvx}, March 2014.

\bibitem[Grygorash et~al.(2006)Grygorash, Zhou, and Jorgensen]{GrygorashEtAl06}
Oleksandr Grygorash, Yan Zhou, and Zach Jorgensen.
\newblock Minimum spanning tree based clustering algorithms.
\newblock In \emph{Proceedings of the 18th IEEE International Conference on Tools with Artificial Intelligence}, page 73–81, 2006.

\bibitem[Hallac et~al.(2015)Hallac, Leskovec, and Boyd]{hallac2015network}
David Hallac, Jure Leskovec, and Stephen Boyd.
\newblock Network lasso: Clustering and optimization in large graphs.
\newblock In \emph{Proceedings of the 21th ACM SIGKDD international conference on knowledge discovery and data mining}, pages 387--396, 2015.

\bibitem[Hastie et~al.(2001)Hastie, Tibshirani, and Friedman]{ESL}
Trevor Hastie, Robert Tibshirani, and Jerome Friedman.
\newblock \emph{The Elements of Statistical Learning}.
\newblock Springer New York Inc., New York, NY, USA, 2001.

\bibitem[Hocking et~al.(2011)Hocking, Joulin, Bach, and Vert]{hocking2011clusterpath}
Toby~Dylan Hocking, Armand Joulin, Francis Bach, and Jean-Philippe Vert.
\newblock Clusterpath an algorithm for clustering using convex fusion penalties.
\newblock In \emph{28th international conference on machine learning}, page~1, 2011.

\bibitem[Johnson(2013)]{johnson2013dynamic}
Nicholas~A Johnson.
\newblock A dynamic programming algorithm for the fused lasso and l 0-segmentation.
\newblock \emph{Journal of Computational and Graphical Statistics}, 22\penalty0 (2):\penalty0 246--260, 2013.

\bibitem[Koklu and Ozkan(2020)]{koklu2020multiclass}
Murat Koklu and Ilker~Ali Ozkan.
\newblock Multiclass classification of dry beans using computer vision and machine learning techniques.
\newblock \emph{Computers and Electronics in Agriculture}, 174:\penalty0 105507, 2020.

\bibitem[Kolmogorov et~al.(2016)Kolmogorov, Pock, and Rolinek]{kolmogorov2016total}
Vladimir Kolmogorov, Thomas Pock, and Michal Rolinek.
\newblock Total variation on a tree.
\newblock \emph{SIAM Journal on Imaging Sciences}, 9\penalty0 (2):\penalty0 605--636, 2016.

\bibitem[Kuthe and Rahmann(2020)]{kuthe2020engineering}
Elias Kuthe and Sven Rahmann.
\newblock Engineering fused lasso solvers on trees.
\newblock In \emph{18th International Symposium on Experimental Algorithms (SEA 2020)}. Schloss Dagstuhl-Leibniz-Zentrum f{\"u}r Informatik, 2020.

\bibitem[Lindsten et~al.(2011)Lindsten, Ohlsson, and Ljung]{lindsten2011clustering}
Fredrik Lindsten, Henrik Ohlsson, and Lennart Ljung.
\newblock Clustering using sum-of-norms regularization: With application to particle filter output computation.
\newblock In \emph{2011 IEEE Statistical Signal Processing Workshop (SSP)}, pages 201--204. IEEE, 2011.

\bibitem[Monath et~al.(2021)Monath, Dubey, Guruganesh, Zaheer, Ahmed, McCallum, Mergen, Najork, Terzihan, Tjanaka, Wang, and Wu]{MonathEtAl21}
Nicholas Monath, Kumar~Avinava Dubey, Guru Guruganesh, Manzil Zaheer, Amr Ahmed, Andrew McCallum, Gokhan Mergen, Marc Najork, Mert Terzihan, Bryon Tjanaka, Yuan Wang, and Yuchen Wu.
\newblock Scalable hierarchical agglomerative clustering.
\newblock In \emph{Proceedings of the 27th ACM SIGKDD Conference on Knowledge Discovery \& Data Mining}, page 1245–1255, 2021.

\bibitem[Murtagh(1983)]{murtagh1983survey}
Fionn Murtagh.
\newblock A survey of recent advances in hierarchical clustering algorithms.
\newblock \emph{The computer journal}, 26\penalty0 (4):\penalty0 354--359, 1983.

\bibitem[Ng et~al.(2001)Ng, Jordan, and Weiss]{ng2001spectral}
Andrew Ng, Michael Jordan, and Yair Weiss.
\newblock On spectral clustering: Analysis and an algorithm.
\newblock \emph{Advances in neural information processing systems}, 14, 2001.

\bibitem[Padilla et~al.(2017)Padilla, Sharpnack, and Scott]{padilla2017dfs}
Oscar Hernan~Madrid Padilla, James Sharpnack, and James~G Scott.
\newblock The dfs fused lasso: Linear-time denoising over general graphs.
\newblock \emph{The Journal of Machine Learning Research}, 18\penalty0 (1):\penalty0 6410--6445, 2017.

\bibitem[Panahi et~al.(2017)Panahi, Dubhashi, Johansson, and Bhattacharyya]{pmlr-v70-panahi17a}
Ashkan Panahi, Devdatt Dubhashi, Fredrik~D. Johansson, and Chiranjib Bhattacharyya.
\newblock Clustering by sum of norms: Stochastic incremental algorithm, convergence and cluster recovery.
\newblock In \emph{Proceedings of the 34th International Conference on Machine Learning}, volume~70 of \emph{Proceedings of Machine Learning Research}, pages 2769--2777. PMLR, 2017.

\bibitem[Pelckmans et~al.(2005)Pelckmans, De~Brabanter, Suykens, and De~Moor]{pelckmans2005convex}
Kristiaan Pelckmans, Joseph De~Brabanter, Johan~AK Suykens, and B~De~Moor.
\newblock Convex clustering shrinkage.
\newblock In \emph{PASCAL Workshop on Statistics and Optimization of Clustering Workshop}, 2005.

\bibitem[Radchenko and Mukherjee(2017)]{radchenko2017convex}
Peter Radchenko and Gourab Mukherjee.
\newblock Convex clustering via l1 fusion penalization.
\newblock \emph{Journal of the Royal Statistical Society: Series B (Statistical Methodology)}, 79\penalty0 (5):\penalty0 1527--1546, 2017.

\bibitem[Reddy et~al.(2011)Reddy, Mishra, and Jana]{reddy2011mst}
Damodar Reddy, Devender Mishra, and Prasanta~K Jana.
\newblock Mst-based cluster initialization for k-means.
\newblock In \emph{International Conference on Computer Science and Information Technology}, pages 329--338. Springer, 2011.

\bibitem[Sun et~al.(2021)Sun, Toh, and Yuan]{sun2021convex}
Defeng Sun, Kim-Chuan Toh, and Yancheng Yuan.
\newblock Convex clustering: Model, theoretical guarantee and efficient algorithm.
\newblock \emph{J. Mach. Learn. Res.}, 22\penalty0 (9):\penalty0 1--32, 2021.

\bibitem[Tan and Witten(2015)]{tan2015statistical}
Kean~Ming Tan and Daniela Witten.
\newblock Statistical properties of convex clustering.
\newblock \emph{Electronic journal of statistics}, 9\penalty0 (2):\penalty0 2324, 2015.

\bibitem[Tibshirani et~al.(2005)Tibshirani, Saunders, Rosset, Zhu, and Knight]{tibshirani2005sparsity}
Robert Tibshirani, Michael Saunders, Saharon Rosset, Ji~Zhu, and Keith Knight.
\newblock Sparsity and smoothness via the fused lasso.
\newblock \emph{Journal of the Royal Statistical Society: Series B (Statistical Methodology)}, 67\penalty0 (1):\penalty0 91--108, 2005.

\bibitem[Tibshirani(2017)]{tibshirani2017dykstra}
Ryan~J Tibshirani.
\newblock Dykstra's algorithm, admm, and coordinate descent: Connections, insights, and extensions.
\newblock \emph{Advances in Neural Information Processing Systems}, 30, 2017.

\bibitem[Tibshirani et~al.(2011)Tibshirani, Taylor, et~al.]{tibshirani2011solution}
Ryan~J Tibshirani, Jonathan Taylor, et~al.
\newblock The solution path of the generalized lasso.
\newblock \emph{The Annals of Statistics}, 39\penalty0 (3):\penalty0 1335--1371, 2011.

\bibitem[Touw et~al.(2022)Touw, Groenen, and Terada]{Daniel:CCMM}
Daniel J.~W. Touw, Patrick J.~F. Groenen, and Yoshikazu Terada.
\newblock Convex clustering through mm: An efficient algorithm to perform hierarchical clustering.
\newblock \emph{arXiv preprint arXiv:2211.01877}, 2022.

\bibitem[Wang et~al.(2018)Wang, Zhang, Sun, and Fang]{wang2018sparse}
Binhuan Wang, Yilong Zhang, Will~Wei Sun, and Yixin Fang.
\newblock Sparse convex clustering.
\newblock \emph{Journal of Computational and Graphical Statistics}, 27\penalty0 (2):\penalty0 393--403, 2018.

\bibitem[Wang et~al.(2023)Wang, Yao, and Allen]{supervised}
Minjie Wang, Tianyi Yao, and Genevera~I. Allen.
\newblock Supervised convex clustering.
\newblock \emph{Biometrics}, n/a\penalty0 (n/a), 2023.
\newblock \doi{https://doi.org/10.1111/biom.13860}.
\newblock URL \url{https://onlinelibrary.wiley.com/doi/abs/10.1111/biom.13860}.

\bibitem[Weylandt et~al.(2020)Weylandt, Nagorski, and Allen]{weylandt2020dynamic}
Michael Weylandt, John Nagorski, and Genevera~I Allen.
\newblock Dynamic visualization and fast computation for convex clustering via algorithmic regularization.
\newblock \emph{Journal of Computational and Graphical Statistics}, 29\penalty0 (1):\penalty0 87--96, 2020.

\bibitem[Wu et~al.(2013)Wu, Li, Jiao, Wang, and Sun]{WuEtAl13}
Jianshe Wu, Xiaoxiao Li, Licheng Jiao, Xiaohua Wang, and Bo~Sun.
\newblock Minimum spanning trees for community detection.
\newblock \emph{Physica A: Statistical Mechanics and its Applications}, 392\penalty0 (9):\penalty0 2265--2277, 2013.

\bibitem[Xu et~al.(2002)Xu, Olman, and Xu]{XuEtAl02}
Ying Xu, Victor Olman, and Dong Xu.
\newblock {Clustering gene expression data using a graph-theoretic approach: an application of minimum spanning trees}.
\newblock \emph{Bioinformatics}, 18\penalty0 (4):\penalty0 536--545, 2002.

\bibitem[Yagishita and Gotoh(2024)]{Non-convexExtension2024}
Shotaro Yagishita and Junya Gotoh.
\newblock Pursuit of the cluster structure of network lasso: Recovery condition and non-convex extension.
\newblock \emph{Journal of Machine Learning Research}, 25\penalty0 (21):\penalty0 1--42, 2024.
\newblock URL \url{http://jmlr.org/papers/v25/21-1190.html}.

\bibitem[Yang et~al.(2016)Yang, Ma, Zhang, Li, and Zhang]{yang2016minimum}
J~Yang, Y~Ma, X~Zhang, S~Li, and Y~Zhang.
\newblock A minimum spanning tree-based method for initializing the k-means clustering algorithm.
\newblock \emph{International Journal of Computer and Information Engineering}, 11\penalty0 (1):\penalty0 13--17, 2016.

\bibitem[Yu et~al.(2015)Yu, Hillebrand, Tewarie, Meier, van Dijk, Van~Mieghem, and Stam]{yu2015hierarchical}
Meichen Yu, Arjan Hillebrand, Prejaas Tewarie, Jil Meier, Bob van Dijk, Piet Van~Mieghem, and Cornelis~Jan Stam.
\newblock Hierarchical clustering in minimum spanning trees.
\newblock \emph{Chaos: An Interdisciplinary Journal of Nonlinear Science}, 25\penalty0 (2):\penalty0 023107, 2015.

\bibitem[Yuan and Lin(2006)]{yuan2006model}
Ming Yuan and Yi~Lin.
\newblock Model selection and estimation in regression with grouped variables.
\newblock \emph{Journal of the Royal Statistical Society Series B: Statistical Methodology}, 68\penalty0 (1):\penalty0 49--67, 2006.

\bibitem[Yuan et~al.(2018)Yuan, Sun, and Toh]{yuan2018efficient}
Yancheng Yuan, Defeng Sun, and Kim-Chuan Toh.
\newblock An efficient semismooth newton based algorithm for convex clustering.
\newblock In \emph{International Conference on Machine Learning}, pages 5718--5726. PMLR, 2018.

\bibitem[Zhang et~al.(2021)Zhang, Chen, and Terada]{zhang2021dynamic}
Bingyuan Zhang, Jie Chen, and Yoshikazu Terada.
\newblock Dynamic visualization for l1 fusion convex clustering in near-linear time.
\newblock In \emph{Uncertainty in Artificial Intelligence}, pages 515--524. PMLR, 2021.

\bibitem[Zhong et~al.(2011)Zhong, Miao, and Fränti]{ZhongEtAl11}
Caiming Zhong, Duoqian Miao, and Pasi Fränti.
\newblock Minimum spanning tree based split-and-merge: A hierarchical clustering method.
\newblock \emph{Information Sciences}, 181\penalty0 (16):\penalty0 3397--3410, 2011.

\bibitem[Zhou et~al.(2021)Zhou, Yi, Mishne, and Chi]{scalable2021}
Weilian Zhou, Haidong Yi, Gal Mishne, and Eric Chi.
\newblock Scalable algorithms for convex clustering.
\newblock In \emph{2021 IEEE Data Science and Learning Workshop (DSLW)}, pages 1--6, 2021.
\newblock \doi{10.1109/DSLW51110.2021.9523411}.

\end{thebibliography}

\newpage
\section*{Supplementary}

The supplementary materials are organized as follows. 
In Section \ref{sec:A1}, we describe the details of the procedure for generating the synthetic data. 
In Section \ref{sec:A2}, we describe the real datasets in evaluation.
In Section \ref{sec:A3}, we discuss a side issue raised by outliers and provide a possible solution to overcome the problem.
In Section \ref{sec:A4}, we discuss the effect of the bandwidth $\gamma$ in the Gaussian kernel on clustering performance. Specifically, we study the effect of the bandwidth $\gamma$ on the performances of the convex clustering methods. 
In Section \ref{sec:A5}, we provide the additional simulation results with different sample sizes $n = 200$ and $5000$ for the convex clustering methods. 
In Section \ref{sec:A6}, we provide the results on the real datasets for hierarchical clustering, non-hierarchical clustering, and convex clustering methods.

\section{Synthetic data generation}
\label{sec:A1}
The synthetic datasets in the numerical experiments are generated as follows:
\begin{itemize}
    \item The Gaussian Mixture 1 (GM1) model generates data points from three two-dimensional Gaussian distributions. To generate $n$ data points, the model independently draws 
    $(\left\lfloor n/3 \right\rfloor, \left\lfloor n/3 \right\rfloor, n - \left\lfloor 2n/3 \right\rfloor)$ data points from three Gaussian distributions, each with distinct mean vectors and a common covariance matrix. The mean vectors are: 
    \begin{equation*}
        \mu_1=(1,2.5)^T, \mu_2=(2.5,-1.8)^T, \mu_3=(-2.5,-2)^T.    
    \end{equation*}
    and the common covariance matrix is the identity matrix $\Sigma = I$.
    \item The Gaussian Mixture 2 (GM2) model generates samples in a manner similar to the GM1 model but with different mean vectors and a distinct covariance matrix. The mean vectors and the common covariance matrices are set as follows:
    \begin{equation*}
    \mu_1 = (1.3, 3.5)^T,\; \mu_2 = (2, -2)^T,\;\mu_3 = (-1.2, 4)^T,\;\text{ and }\;
    \Sigma = \begin{pmatrix} 1 &0.9 \\ 0.9 &1.2\end{pmatrix}.    
    \end{equation*}
    \item The Two Moons (TM) model consists of two half moons, each with an equal sample size $(n/2, n/2)$. The x-coordinates of the first half moon are generated uniformly from $[0,\pi]$, with corresponding y-coordinates given by $y = 2\sin(x)-0.35$. For the second half moon, the x-coordinates are generated uniformly from $[\pi/2,3\pi/2]$, and the y-coordinates are given by $y = 2\cos(x)-0.35$. 
    Finally, Gaussian noise $\mathcal{N}(0, 0.25^2)$ is independently added to the x and y coordinates of each data point.
    \item The Two Circles (TC) model consists of two concentric circles with the same center but different radii. Each circle has the same number of data points. For each circle, the coordinates are defined as $x = t\times \sin(2\pi l), y = t\times \cos(2\pi l)$, where $l$ is uniformly generated from $[0,1]$ and $t$ represents the given radius. For the outer circle, 9/10 of the data points are generated with $t\in [0.8, 0.9]$ and 1/10 points with $t\in [0.6,0.8]$. For the inner circle, 9/10 of the data points are generated with $t\in [0.3, 0.5]$ and 1/10 with $t\in [0.4, 0.6]$.
    \item The Multidimensional Three Cluster (MTC) model generates samples from a multivariate Gaussian distribution with three different mean vectors and random covariance matrices. The three $p$-dimensional mean vectors are created by first generating three two-dimensional vectors:
    \begin{equation*}
        \begin{aligned}
            \mu_1 = (4/\sqrt{3}, 0), \, \mu_2 = (-2/\sqrt{3}, 2), \, \mu_3 = (-2/\sqrt{3}, -2)
        \end{aligned}
    \end{equation*}
    Then, these two-dimensional vectors are multiplied by a random orthogonal rotation matrix $A \in \mathbb{R}^{p \times 2}$ such that $AA^T = \boldsymbol{I}_p$, and added to a $p$-dimensional vector $\nu_i$:
    \begin{equation*}
        \mu_i A^T + \nu_i 
    \end{equation*}
    whose elements are generated from $\mathcal{N}(0, 0.25^2)$. 
    The three covariance matrices are generated similarly. 
    First, a covariance matrix $\Sigma_B$ is calculated from a randomly generated matrix $B \in \mathbb{R}^{50 \times p}$, 
    where $B_{ij} \sim \mathcal{N}(0, 1)$ for $i = 1, \dots, 50$ and $j = 1, \dots, p$. 
    The final covariance matrices are set to be $\Sigma = \Sigma_B \otimes ll^T$, where $l \in \mathbb{R}^p$ is a $p$-dimensional vector with elements uniformly drawn from the interval $[0.5, 1.4]$.
    \item The checkerboard model follows the setting described in \citep{chi2017convex}. 
    The checkerboard model generates a bi-cluster-structured square data matrix $X$, 
    where its element $x_{ij} \overset{i.i.d}{\sim} \, \mathcal{N}(\mu_{rc}, \sigma^2)$, 
    and $\mu_{rc}$ takes one of the values equally spaced by 0.5 between -6 and 6. 
    A low signal-to-noise ratio setting is considered, where the standard deviation is set to $\sigma=3$. 
    We consider 4 row groups and 4 column clusters. 
    The resulting data matrix consists of 16 bi-clusters. The data points are generated from one of the row groups randomly, and the probability of a row assigned to the $i$-th group is inversely proportional to $i$. Similarly, the column groups are assigned in the same manner.
    \item The Four Spherical (FS) model follows the setting described in \citep{wang2018sparse}. 
    The FS model generates samples from 4 multivariate Gaussian distributions (MVN). 
    Each data point is randomly sampled from one of the MVNs, and the first 20 informative features are generated from $\mathbf{MVN}(\boldsymbol{\mu}_k, \boldsymbol{I}_{20})$, $k \in \{1,2,3,4\}$. The mean vectors, $\bm{\mu}_k$ are defined as follows:
    \begin{equation*}
    \begin{aligned}
        \bm{\mu}_k &= 
        (\mu\bm{1}_{10}^T, \mu\bm{1}_{10}^T) I(k=1) + 
        (\mu\bm{1}_{10}^T, -\mu\bm{1}_{10}^T) I(k=2) + \\
        &
        (-\mu\bm{1}_{10}^T, \mu\bm{1}_{10}^T) I(k=3) + 
        (-\mu\bm{1}_{10}^T, -\mu\bm{1}_{10}^T) I(k=4) 
    \end{aligned}
    \end{equation*}
    Here, $\mu = 1.5$ is used. Finally, the remaining $n-20$ noisy features are generated independently from $\mathcal{N}(0, 1)$.
\end{itemize}

\section{Real data}
\label{sec:A2}
We summarize the real datasets that are used in the paper. All datasets are available from the UCI repository of the Machine Learning Database \citep{Dua:2019}. 

\begin{description}
\item[\href{https://archive.ics.uci.edu/dataset/109/wine}{Wine}:] A dataset includes 178 samples and 13 features on a chemical analysis of different wines. The dataset has three types of wines: $n_1 = 59$, $n_2 = 71$ and $n_3 = 48$. 
\item[\href{https://archive.ics.uci.edu/dataset/15/breast+cancer+wisconsin+original}{Breast Cancer}:] A dataset includes 449 complete samples and 9 features. Duplicated and incomplete data points are removed. 
The dataset has two types of labels: benign ($n_1=213$) and malignant ($n_2=236$).
\item[\href{https://archive.ics.uci.edu/dataset/50/image+segmentation}{Segmentation}:] A image segmentation dataset from 7 outdoor images. Each data point is a region ($3\times 3$ pixels) extracted from one of the images. The corresponding image is assigned as the class labels, and each label has a balanced number of data points. 
\item[\href{https://blog.nus.edu.sg/mattohkc/softwares/convexclustering/}{Mnist}:] The processed subset of MNIST dataset from SSNAL paper \citep{sun2021convex}, which is available online. It is a subset ($n=10000$) of the MNIST dataset, and the feature dimension is reduced to $p=10$. Each class in the MNIST dataset has almost the same number of data points.
\item[\href{https://archive.ics.uci.edu/dataset/81/pen+based+recognition+of+handwritten+digits}{Pendigits}:] A dataset includes 16 features and 10 labels. The Pendigits dataset is created by collecting handwritten digits from different writers and the sample size $n = 10992$. Each class has almost the same number of points.
\item[\href{https://archive.ics.uci.edu/dataset/602/dry+bean+dataset}{Drybean}:] A dataset includes 16 features such as form and shape of the bean \citep{koklu2020multiclass} and seven bean classes: BARBUNYA ($n_1=1322$), BOMBAY ($n_2=522$), CALI ($n_3=1630$), DERMASON ($n_4=3546$), HOROZ ($n_5=1928$), SEKER ($n_6=2027$), SIRA ($n_7=2636$). 
\item[\href{https://archive.ics.uci.edu/dataset/347/hepmass}{HEPMASS}:] A subset of the HEPMASS dataset used in the main paper. HEPMASS is a collection of signals in high-energy physics experiments. 
HEPMASS has two labels: background noises ($n_1=50130$) or true signals ($n_2=49870$).
\item[\href{https://archive.ics.uci.edu/dataset/31/covertype}{Covtype}:] A forest cover type dataset includes 581012 samples and 54 features. In the dataset, there are 7 different cover types: 1 ($n = 211840$), 2 ($n=283301$), 3 ($n=35754$), 4 ($n=2747$), 5 ($n=9493$), 6 ($n=17367$), 7 ($n=20510$).
\end{description}

\section{On effects of outliers}
\label{sec:A3}
When assigning weights on edges $(i,j)\in E$ using a similarity measure such as the Gaussian kernel, 
edges of outliers could have quite small edge weights. 
Such negligible small weights cause a problem as the outliers are difficult to merge into clusters.
This problem existed not only for TGCC but also for general convex clustering methods, which can be an advantage or a disadvantage.
For example, when we want to detect an outlier as a single cluster, this is a desirable property.
\begin{figure}[ht]
\centering
    \includegraphics[width = 0.65\textwidth]{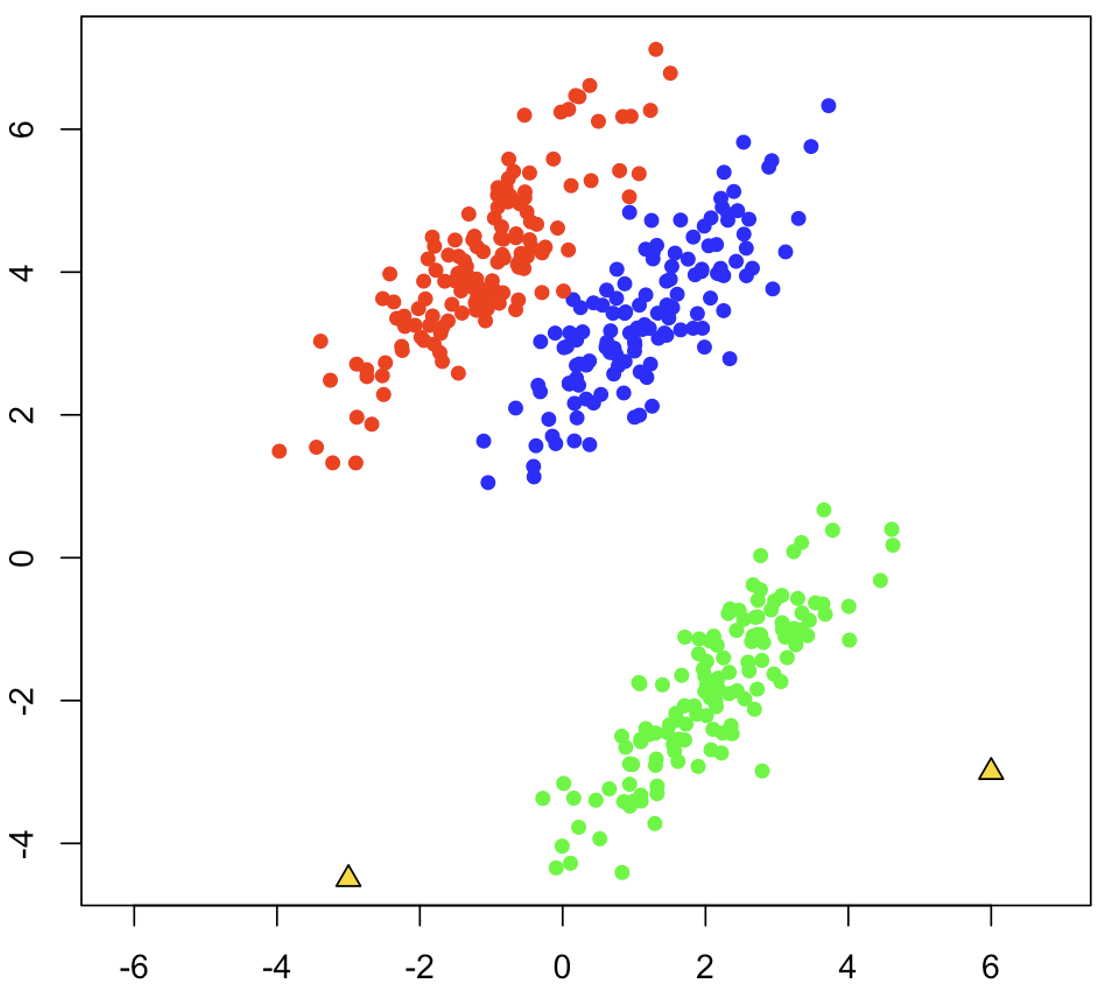} 
    \caption{The Gaussian Mixture 2 (GM2) example with two outliers (denoted by triangle marks).}
    \label{fig:threshold-points}
\end{figure}
On the other hand, when we want to assign outliers to the nearest clusters, 
weights that are too small make it difficult and could induce unstable results.
When outliers exist and have corresponding small edge weights, all points are not merged into the whole cluster, even for large values of $\lambda$. A possible solution is to set a global threshold $\delta$ of weights, i.e., $w'_{ij}(\delta)= \max\{w_{ij}, \delta\}$ for $(i,j)\in E$. The value of $\delta$ can be set as the $10$th quantile point of weights $w_{ij},\forall (i,j)\in E$. In the case of TGCC, the method is based on the tree-structured weight, and we consider the outliers to be mostly likely to be close to the ``leaves'' of the tree structure. Thus we can set the threshold on a subset of nodes whose depths to the closet leaf node are less than a given number $\tilde{D}$. In the paper, we set the $\tilde{D} = 50$ for TGCC. As a result, if the sample size is small, all the nodes may become the target of the threshold. If the sample size is large, we only set the threshold on those nodes close to the leaves. 

\begin{figure}[ht]
\centering
\includegraphics[width = 0.85\textwidth]{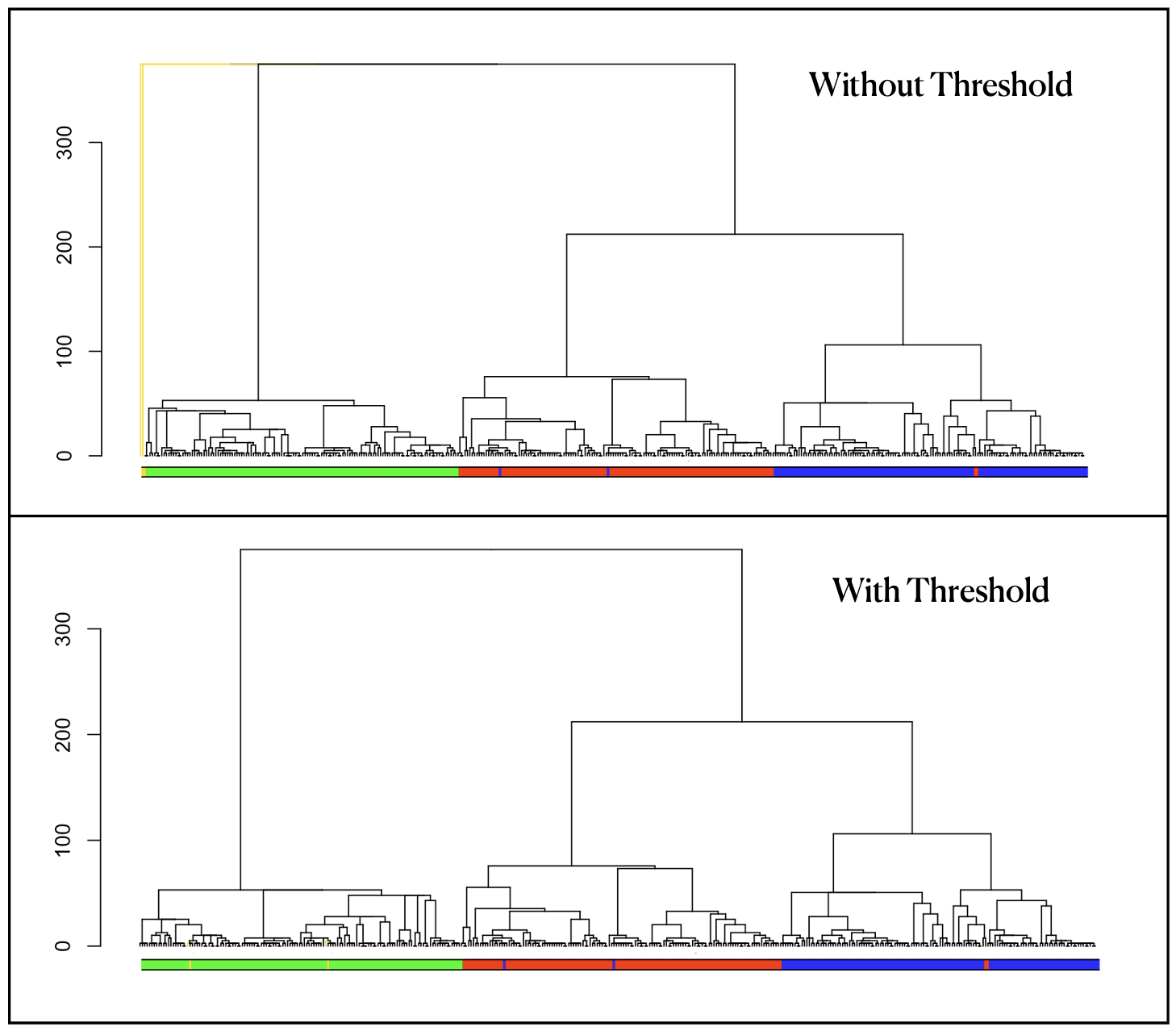} 
\caption{Effect of the threshold in TGCC 
(Top: Dendrogram of TGCC without a threshold; Bottom: Dendrogram of TGCC with a threshold.) The color bars represent the true labels.}
\label{fig:threshold}
\end{figure}

Here, we showed an illustrative example of the effect of outliers. In Figure~\ref{fig:threshold-points}, two outliers were denoted by triangle marks. We applied TGCC with/without the threshold for the dataset (Figure~\ref{fig:threshold-points}). The top panel of Figure~\ref{fig:threshold} showed the dendrogram of TGCC without any threshold.
We detected two outliers easily from this dendrogram.
The bottom panel of Figure~\ref{fig:threshold} showed the dendrogram of TGCC with the threshold.
The dendrogram did not suffer any effect of outliers. The threshold forced the merger of outliers with neighbor clusters.
Both results could be useful in practical situations.

In the numerical experiments,
since we did not want to detect outliers and suffer the effects of outliers, we applied the threshold approach described above for TGCC.

\section{On effects of hyperparameter}
\label{sec:A4}
We discuss the effect of hyperparameter $\gamma$ in the Gaussian kernel. Including but not limited to the convex clustering, the choice of kernel bandwidth $\gamma$ is essentially challenging to decide. Usually, we need to choose $\gamma$ in a data-driven fashion. As we will discuss later, even for samples from the same population, the sample size $n$ can greatly affect the choice of $\gamma$. 

\begin{figure}[ht]
\centering
\includegraphics[width=0.9\linewidth]{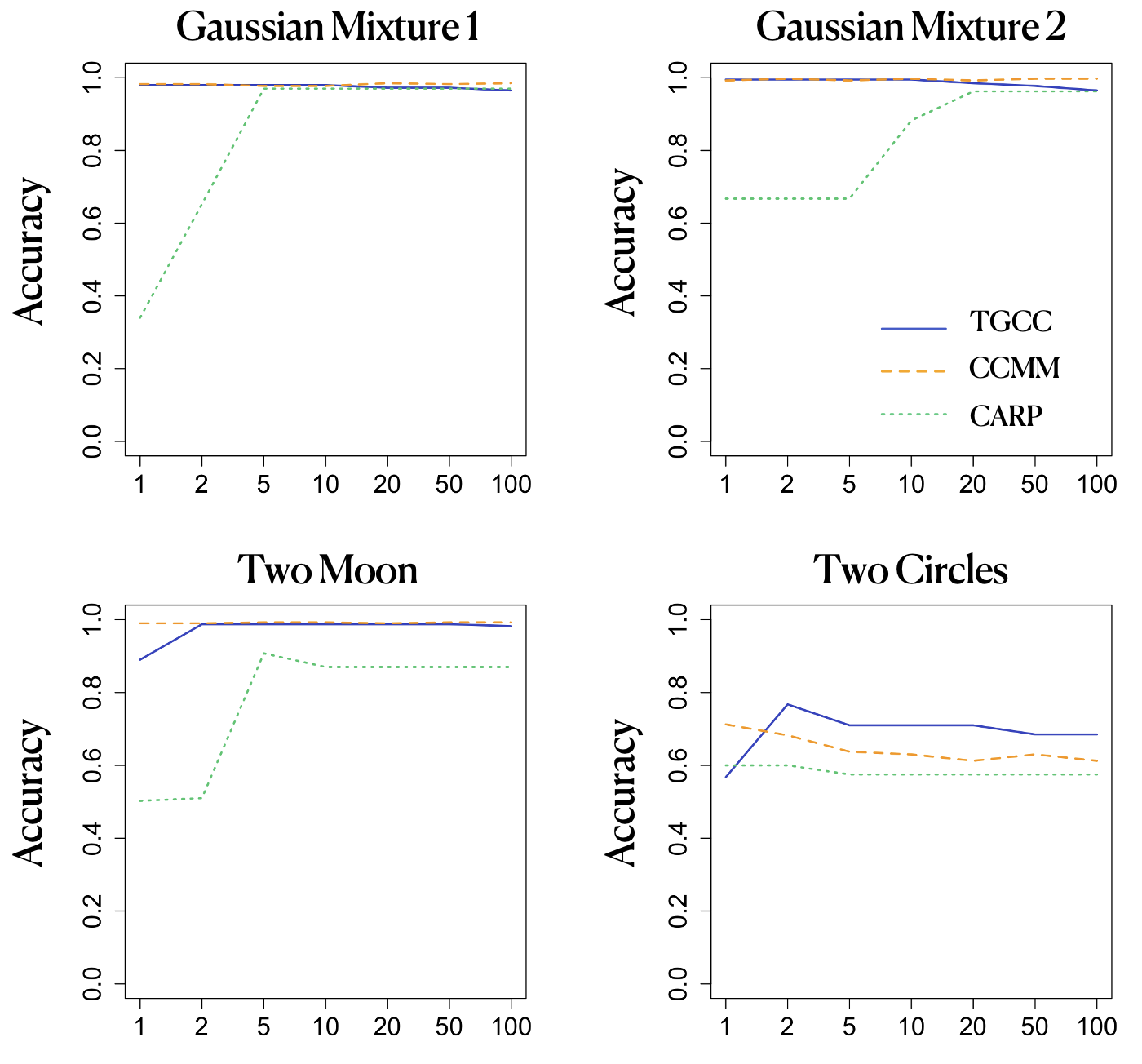}
\caption{Clustering performance on different hyperparameters (Green: SSNAL, Orange: CCMM, Blue: CARP, Red: TGCC).}
\label{fig:Choose_Gamma}
\end{figure}

We studied the effect of $\gamma$ for the convex clustering methods (CARP, CCMM, and TGCC). The sample size was fixed to be $n = 400$ for the four synthetic examples. We considered a shared set of $\gamma \in \{1,2,5,10,20,50,100\}$. Figure \ref{fig:Choose_Gamma} included four figures on each synthetic example. The X-axises showed the different choices of $\gamma$. The Y-axises showed the corresponding clustering accuracy for the convex clustering methods. All the reported numbers were the median values over five repetitions. 

We observed that the optimal values of $\gamma$ vary across different methods. 
The structures of weights are different: tree for TGCC and $k$-NN graph for CARP, SSNAL, and CCMM. Therefore, the values of $\gamma$ cannot be compared directly across methods.  
The results also suggest the selection of $\gamma$ should be chosen in a data-driven fashion. In practice, finding a universally applicable $\gamma$ is challenging, however, techniques such as the normalization method discussed in Section \ref{sec:A5} can be employed to alleviate the difficulty of choosing a proper gamma that caused dataset scales.

\section{Experiments on synthetic datasets}
\label{sec:A5}

We showed the results on synthetic datasets with different sample sizes: $n = 200, 5000$.  In the experiments, as the sample size $n$ grew, data points became more crowded. 
The edges in the graph become shorter, which affects the choice of $\gamma$. 
For example, the distributions of edge distances for different sample sizes $n = 100$ and $n = 10000$ are different. Different scales of the pairwise distances affect the choice of $\gamma$. In order to let the choice of $\gamma$ be independent of the scale of distances on edges $(i,j)\in E$, 
we consider normalizing the weights using $\kappa_n$: 
\[
\kappa_n^2 := \frac{1}{\#(E)}\sum_{(i,j)\in E}\|x_i-x_j\|^2_2,
\]
and assign the weights as follows:
\[
\tilde{w}_{ij}^{(n)}(\gamma)
:=
\begin{cases}
\exp\left(-\dfrac{\|x_i-x_j\|_2^2}{\gamma\kappa_n^2} \right) & \text{if } (i,j)\in E\\
0 & \text{if } (i,j)\not\in E.
\end{cases}
\]
By normalizing the pairwise distances, we no longer need to adjust the scale of $\gamma$ regarding the scale of pairwise distances. 

In Table \ref{table:sim200} and Table \ref{table:sim5000}, classification accuracy (AC) and adjusted random index (ARI) were used to evaluate the clustering performance. We reported the median values of ARI and AC of different methods over 50 repetitions. 

\begin{table}[ht]
    \caption{Clustering performance ($n = 200$). 
    Classification accuracy (AC) and adjusted random index (ARI) are reported. All numbers are the median values in 50 repetitions.
    }  
    \label{table:sim200}
\centering
\begin{sc}
    \begin{tabular}{lccccc}
    \toprule
         Data  & Index & CPAINT  & CARP  & CCMM  & TGCC \\  
    \midrule
         GM1   & AC    & 0.738   & 0.985  & 0.990  & 0.980 \\
               & ARI   & 0.478   & 0.955  & 0.970  & 0.940 \\
    \midrule
         GM2   & AC    & 0.700   & 0.985  & 0.995  & 0.995 \\
               & ARI   & 0.306   & 0.955  & 0.985  & 0.985 \\
    \midrule
         TM    & AC    & 0.685   & 0.923  & 0.990  & 0.973 \\
               & ARI   & 0.036   & 0.713  & 0.960  & 0.893 \\
    \midrule
         TC    & AC    & 0.740   & 0.665  & 0.705  & 0.700 \\
               & ARI   & 0.148   & 0.106  & 0.298  & 0.157 \\
    \bottomrule
\end{tabular}
\end{sc}
\end{table}

\begin{table}[ht]
    \caption{Clustering performance ($n = 5000$). Classification accuracy (AC) and adjusted random index (ARI) are reported. All numbers are the median values in 50 repetitions.}  
    \label{table:sim5000}
\centering
\begin{sc}
    \begin{tabular}{lcccc}
    \toprule
         Data  & Index & CPAINT  & CCMM  & TGCC \\  
    \midrule
         GM1   & AC    & 0.742   & 0.988  & 0.984 \\
               & ARI   & 0.539   & 0.963  & 0.952 \\
    \midrule
         GM2   & AC    & 0.679   & 0.996  & 0.995 \\
               & ARI   & 0.302   & 0.989  & 0.984 \\
    \midrule
         TM    & AC    & 0.664   & 0.994  & 0.991 \\
               & ARI   & 0.001   & 0.975  & 0.966 \\
    \midrule
         TC    & AC    & 0.721   & 0.628  & 0.746 \\
               & ARI   & 0.144   & 0.560  & 0.241 \\
    \bottomrule
\end{tabular}
\end{sc}
\end{table}

For CAPR, CCMM, and TGCC, we chose the best $\gamma$ from a set of candidates that gave the best classification accuracy on the same 50 repetitions. 
The candidate sets of $\gamma$ were prepared for the kNN-based convex clustering methods ($\{0.5,1,2,5,10,50,100\}$) and TGCC ($\{1,2,5,10,20,50,100\}$). We used different candidate sets because the kNN graph is denser than a tree, and to get the centroids $\hat{\boldsymbol{\theta}}$ merged, TGCC needs to assign large weights to edges. 

Although CARP, CCMM, and TGCC produced complete dendrograms, it is not guaranteed to find the exact given number of clusters because multiple clusters may be fused at the same $\lambda$ in convex clustering. We cut the dendrogram where the obtained number of clusters is the closest to the desired number of clusters.
CAPR exceeded the capacity of memory when $n = 5000$. For this reason, we removed CAPR from Table \ref{table:sim5000}.  
We observed that TGCC consistently showed comparable performance to other methods. 

\section{Experiments on real datasets}
\label{sec:A6}
We provide the results on real datasets for three categories of clustering methods: hierarchical clustering (single linkage, complete linkage), non-hierarchical clustering (spectral clustering (SC), $k$-means, DBSCAN), and convex clustering methods (CPAINT, CARP, CCMM, TGCC). 

We show the results on the six datasets: Wine, Breast Cancer, Segmentation, Mnist, Pendigits, and Drybeans. For most hierarchical clustering methods, to improve computational efficiency, they need to store a pre-computed distance matrix to avoid computing the distance between sample pairs for more than once. However, even a limited sample size $n = 10^5$ can become infeasible for a personal computer \citep{gagolewski2016genie}. DBSCAN, Spectral clustering and CARP methods are computationally infeasible for large-scale datasets. 
Thus, we reported their result on the real datasets whose sample sizes $n \le 100000$. All runtimes are the mean over 3 independent repetitions. 


\begin{table}[ht]
    \caption{Clustering performance on real datasets. Classification accuracy (AC) and adjusted random index (ARI) are reported.}  
    \label{table:realcomplete}
\centering
\resizebox{0.9\columnwidth}{!}{%
    \begin{tabular}{lc K{1.2cm}K{1.2cm}K{1.2cm} K{1.2cm}K{1.2cm}K{1.3cm}K{1.4cm}K{1.4cm}}
        \toprule
        Method   & Index     & Wine  & Cancer & Seg.  & MNIST & Pen.  & Dry.   & HEPM. & Covt.\\
                 & \it{n}    & 178   & 449    & 2310  & 10000 & 10992 & 13611  & 100000  & 581012  \\
                 & \it{p}    & 13    & 9      & 18    & 10    & 16    & 16     & 5       & 11 \\
        \midrule
        Single   & AC        & 0.399  & 0.526  & 0.146  & 0.115  & 0.111  & 0.261  & -- & -- \\
                 & ARI       & -0.007 & 0.000  & 0.000  & 0.000  & 0.000  & 0.000  & -- & -- \\
                 & Run time  & 0.000  & 0.003  & 0.097  & 1.840  & 2.441  & 3.605  & -- & -- \\
        \midrule
        Complete & AC        & 0.837  & 0.913  & 0.167  & 0.918  & 0.452  & 0.505  & -- & -- \\
                 & ARI       & 0.577  & 0.682  & 0.001  & 0.830  & 0.249  & 0.362  & -- & -- \\
                 & Run time  & 0.001  & 0.004  & 0.099  & 1.742  & 2.475  & 3.881  & -- & -- \\
        \midrule
        Spectral & AC        & 0.955  & 0.955  & 0.519  & 0.921  & 0.893  & 0.807  & -- & -- \\
                 & ARI       & 0.865  & 0.829  & 0.318  & 0.835  & 0.800  & 0.645  & -- & -- \\
                 & Run time  & 0.097  & 0.234  & 1.512  & 11.430 & 12.176 & 16.155 & -- & -- \\
        \midrule
        DBSCAN   & AC        & 0.399  & 0.933  & 0.610  & 0.815  & 0.600  & 0.305  & -- & -- \\
                 & ARI       & 0.000  & 0.750  & 0.308  & 0.699  & 0.194  & 0.079  & -- & -- \\
                 & Run time  & 0.000  & 0.002  & 0.058  & 0.697  & 1.118  & 1.824  & -- & -- \\
        \midrule
        $k$-means & AC       & 0.966  & 0.929  & 0.555  & 0.920  & 0.715  & 0.798  & 0.880 & 0.259 \\
                 & ARI       & 0.897  & 0.735  & 0.461  & 0.834  & 0.552  & 0.669  & 0.577 & 0.020 \\ 
                 & Run time  & 0.029  & 0.030  & 0.422  & 1.741  & 4.175  & 3.100  & 4.228 & 218.977 \\ 
        \midrule
        CARP     & AC        & 0.607  & 0.949  &  --    & --     & --     & --     & -- & -- \\
                 & ARI       & 0.024  & 0.805  &  --    & --     & --     & --     & -- & -- \\
                 & Run time  & 0.112  & 4.295  &  --    & --     & --     & --     & -- & -- \\
        \midrule
        CPAINT   & AC        & 0.657  & 0.679  & 0.218  & 0.403  & 0.385  & 0.577  & -- & -- \\
                 & ARI       & 0.478  & 0.170  & 0.006  & 0.050  & 0.099  & 0.251  & -- & -- \\
                 & Run time  & 0.829  & 0.131  & 1.204  & 3.300  & 7.076  & 22.716  & -- & -- \\
        \midrule
        CCMM     & AC        & 0.972  & 0.949  & 0.581  & 0.877  & 0.714  & 0.551  & 0.533 & -- \\
                 & ARI       & 0.914  & 0.805  & 0.310  & 0.805  & 0.694  & 0.402  & 0.424 & -- \\
                 & Run time  & 0.089  & 0.262  & 2.824  & 18.134 & 37.386 & 50.664 & 159.809 & -- \\
        \midrule
        TGCC     & AC        & 0.910  & 0.920  & 0.535  & 0.918  & 0.840  & 0.737  & 0.892 & 0.488 \\
                 & ARI       & 0.741  & 0.704  & 0.388  & 0.831  & 0.762  & 0.576  & 0.617 & 0.003 \\
                 & Run time  & 0.109  & 0.090  & 0.629  & 0.855  & 1.619  & 1.768  & 5.790 & 264.148 \\
        \bottomrule
\end{tabular}
}
\end{table}


\end{document}